\documentclass[10pt,journal,compsoc,draftcls]{IEEEtran}
\usepackage{color,soul}
\sethlcolor{yellow}
\renewcommand\hl[1]{#1}

\usepackage[framemethod=tikz]{mdframed}

\makeatletter 
\newcount\SOUL@minus
\makeatother  

\ifCLASSOPTIONcompsoc
  \usepackage[nocompress]{cite}
\else
  \usepackage{cite}
\fi
\ifCLASSINFOpdf
\else
\fi
\usepackage{hyperref}

\usepackage{array}
\usepackage{ragged2e}
\newcolumntype{L}[1]{>{\raggedright\let\newline\\\arraybackslash\hspace{0pt}}m{#1}}
\newcolumntype{C}[1]{>{\centering\let\newline\\\arraybackslash\hspace{0pt}}m{#1}}
\newcolumntype{R}[1]{>{\raggedleft\let\newline\\\arraybackslash\hspace{0pt}}m{#1}}

\usepackage{subcaption}
\usepackage{multirow}
\usepackage{amssymb, amsmath}
\usepackage{longtable}
\usepackage{listings}
\usepackage[ruled,vlined]{algorithm2e}

\usepackage{color, colortbl}
\definecolor{Gray}{gray}{0.9}
\definecolor{LightCyan}{rgb}{0.88,1,1}
\definecolor{Cyan}{rgb}{0.5,1,1}
\definecolor{Orange}{rgb}{1,0.8,0}
\definecolor{Yellow}{rgb}{1,1,0}
\definecolor{Green}{rgb}{0.5,0.9,0}
\definecolor{Blue}{rgb}{0,0.7,1}
\definecolor{LtBlue}{rgb}{0.4,0.8,1}
\definecolor{Purple}{rgb}{.7,0.5,1}

\usepackage{hyperref}

\usepackage[disable]{todonotes}

\usepackage{float}

\newcommand{\isabelle}[1]{\todo[inline,color=orange!40]{#1 -- Isabelle}} 

\newcommand{\zhengying}[1]{\todo[inline,color=yellow!40]{#1 -- Zhengying}}
\newcommand{\adrien}[1]{\todo[inline,color=green!40]{#1 -- Adrien}}
\definecolor{turquoise}{RGB}{64,224,208}

\newcommand{\julio}[1]{\todo[inline,color=magenta]{#1 -- Julio}}

 \newcommand{\michael}[1]{\todo[inline,color=orange!40]{#1 -- Michael}}

\newcommand{\participant}[1]{\textit{#1}}
\newcommand{\task}[1]{\textit{#1}}

\newcounter{magicrownumbers}
\newcommand\rownumber{\stepcounter{magicrownumbers}\arabic{magicrownumbers}}

\usepackage[switch]{lineno}

\begin{document}


%
\title{Winning solutions and post-challenge analyses of the ChaLearn AutoDL challenge 2019}
%
%
%
%

\author{
Zhengying Liu*,
Adrien Pavao*,
Zhen Xu*,
Sergio Escalera, Fabio Ferreira, Isabelle Guyon,  Sirui Hong, Frank Hutter, Rongrong~Ji, Julio C. S. Jacques Junior, Ge Li,  Marius Lindauer, Zhipeng Luo, Meysam Madadi, Thomas Nierhoff,  Kangning Niu, Chunguang Pan, Danny~Stoll, Sebastien Treguer, Jin Wang, Peng Wang, Chenglin Wu, Youcheng Xiong, Arb\"{e}r Zela, Yang Zhang


\thanks{*The first three authors contributed equally. The other authors are in alphabetical order of last name. The corresponding author is: Zhengying Liu (\textit{zhengying.liu@inria.fr}), with Université Paris-Saclay. 
}.
}

\IEEEtitleabstractindextext{%

\begin{abstract} 
\hl{
This paper reports the results and post-challenge analyses of ChaLearn's AutoDL challenge series, which helped sorting out a profusion of AutoML solutions for Deep Learning (DL) that had been introduced in a variety of settings, but lacked fair comparisons. 
All input data modalities (time series, images, videos, text, tabular) were formatted as tensors and all tasks were multi-label classification problems. Code submissions were executed on hidden tasks, with limited time and computational resources, pushing solutions that get results quickly.
In this setting, DL methods dominated, though popular Neural Architecture Search (NAS) was impractical. Solutions relied on fine-tuned pre-trained networks, with architectures matching data modality. Post-challenge tests did not reveal improvements beyond the imposed time limit. While no component is particularly original or novel, a high level modular organization emerged featuring a ``meta-learner'', ``data ingestor'', ``model selector'', ``model/learner'', and ``evaluator''. This modularity enabled ablation studies, which revealed the importance of (off-platform) meta-learning, ensembling, and efficient data management. Experiments on heterogeneous module combinations further confirm the (local) optimality of the winning solutions. Our challenge legacy includes an ever-lasting benchmark }(\url{http://autodl.chalearn.org}), \hl{the open-sourced code of the winners, and a free ``AutoDL self-service''.
}
\end{abstract}

\begin{IEEEkeywords}
AutoML, Deep Learning, Meta-learning, Neural Architecture Search, Model Selection, Hyperparameter Optimization
\end{IEEEkeywords}}

\maketitle

\IEEEdisplaynontitleabstractindextext

%
\IEEEpeerreviewmaketitle


%
%
%
%


 

\ifCLASSOPTIONpeerreview
\begin{center} \bfseries EDICS Category: 3-BBND \end{center}
\fi
%
\IEEEpeerreviewmaketitle


t\IEEEraisesectionheading{\section{Introduction} \label{intro}}

\IEEEPARstart{T}{he} \hl{AutoML problem asks whether one could have a single algorithm} (an \emph{AutoML algorithm}) \hl{that can perform learning on a large spectrum of tasks with consistently good performance, removing the need for human expertise (in defiance of ``No Free Lunch'' theorems} {\cite{wolpert_no_1997, wolpert_lack_1996, wolpert_supervised_2001}}). \hl{
Our goal is to evaluate and foster the improvement of methods that solve the AutoML problem, emphasizing Deep Learning approaches. To that end, we organized in 2019 the  Automated Deep Learning (AutoDL) challenge series}~\cite{liu_autodl_2019}, \hl{which provides a reusable benchmark suite. Such challenges encompass a variety of domains in which Deep Learning has been successful: computer vision, natural language processing, speech recognition, as well as classic tabular data (feature-vector representation).} 

AutoML is crucial to accelerate data science and reduce the need for data scientists and machine learning experts. For this reason, much effort has been drawn towards achieving true AutoML, both in academia and the private sector. In academia, AutoML challenges \cite{guyon_analysis_2018} have been organized and collocated with top machine learning conferences such as ICML and NeurIPS to motivate AutoML research in the machine learning community. The winning approaches from such prior challenges (e.g. auto-sklearn \cite{feurer_efficient_2015}) are now widely used both in research and in industry. More recently, interest in Neural Architecture Search (NAS) has exploded~\cite{ elsken_neural_2019,baker_designing_2017, negrinho_deeparchitect:_2017, cai_proxylessnas:_2018, liu_darts_2019}. On the industry side, many companies such as Microsoft \cite{fusi_probabilistic_2018} and Google are developing AutoML solutions. Google has also launched various AutoML \cite{cortes_adanet_2017}, NAS \cite{zoph_neural_2016, real_large-scale_2017, pham_efficient_2018, real_automl-zero_2020}, and meta-learning \cite{finn_model-agnostic_2017, finn_online_2019} research efforts. Most of the above approaches, especially those relying on Hyper-Parameter Optimization (HPO) or NAS, require significant computational resources and engineering time to find good models. Additionally, reproducibility is impaired by undocumented heuristics \cite{yang_nas_2020}. 
\hl{
Drawn by the aforementioned great potential of AutoML in both academia and industry, a collaboration led by ChaLearn, Google and 4Paradigm was launched in 2018 and a competition in AutoML applied to Deep Learning was conceived, which was the inception of the AutoDL challenge series. To our knowledge, this was the first machine learning competition (series) ever soliciting AutoDL solutions. In the course of the design and implementation we had to overcome many difficulties. We made extensive explorations and revised our initial plans, leading us to organize a series of challenges rather than a single one. 
In this process, we formatted 66 datasets constituting a reusable benchmark resource. Our data repository is still growing, as we continue organizing challenges on other aspects of AutoML, such as the recent AutoGraph competition. 
In terms of competition protocol, our design provides a valuable example of a system that evaluate AutoML solutions, with features such as (1) multiple tasks execution aggregated with average rank metric; (2) emphasis on any-time learning that urges trade-off between accuracy and learning speed; (3) separation of feedback phase and final blind test phase that prevents leaderboard over-fitting. Our long-lasting effort in preparing and running challenges for 2 years is harvested in this paper, which analyses more particularly the last challenge in the series (simply called AutoDL), which featured} {\bf datasets from heterogeneous domains}, \hl{as opposed to previous challenges that were domain specific.

The AutoDL challenge analysed in this paper is the culmination of the AutoDL challenge series, whose major motivation} is two-fold. First, we desire to continue promoting the community's research interests on AutoML to build universal AutoML solutions that can be applied to any task (as long as the data is collected and formatted in the same manner). By choosing tasks in which Deep Learning methods excel, we put gentle pressure on the community to improve on Automated Deep Learning. Secondly, we create a reusable benchmark for fairly evaluating AutoML approaches, on a wide range of domains. Since computational resources and time cost can be a non-negligible factor, we introduce an \emph{any-time learning metric} called Area under Learning Curve (ALC) (see Section \ref{subsec:metric}) for the evaluation of participants' approaches, taking into consideration both the final performance (e.g. accuracy) and the speed to achieve this performance (using wall-time). As far as we know, the AutoDL challenges are the only competitions that adopt a similar any-time learning metric.

Acknowledging the difficulty of engineering universal AutoML solutions, we first organized four preliminary challenges. Each of them focused on a specific application domain. These included: AutoCV for images, AutoCV2 for images and videos, AutoNLP for natural language processing (NLP) and AutoSpeech for speech recognition. Then, during NeurIPS 2019 we launched the final AutoDL challenge that combined all these application domains, and tabular data. All these challenges shared the same competition protocol and evaluation metric ({\em i.e.} ALC) and provided data in a similar format. All tasks were multi-label classification problems. 


For domain-specific challenges such as AutoCV, AutoCV2, AutoNLP and AutoSpeech, the challenge results and analysis are presented in \cite{liu_autodl_2019} \hl{and some basic information can be found in Table} \ref{tab:challenge-stats}. 
\begin{table}[t]
    \caption{\textbf{Basic facts on AutoDL challenges}.
    } 
    \centering
    \setlength\tabcolsep{4pt}
    \scriptsize
    \begin{tabular}{|c|c|c|c|c|c|c|c|c|}
    \hline
         Challenge  & Begin date & End date & \#Teams & \#Submis- & \#Phases 
         \\
            & 2019  &  2019-20 &   & sions   &  
         \\ \hline \hline
         AutoCV1  & May 1 & Jun 29 & $102$ & $938$ & 1 
         \\ \hline
         AutoCV2  & Jul 2 & Aug 20 & 34 & 336 & 2 
         \\ \hline
         AutoNLP  & Aug 2 & Aug 31 & 66 & 420 & 2 
         \\ \hline
         AutoSpeech  & Sep 16 & Oct 16 & 33 & 234 & 2 
         \\ \hline
         {\textbf AutoDL}  & \textbf{Dec 14} & \textbf{Apr 3} & \textbf{28} & \textbf{80} & \textbf{2} 
         \\ \hline
         
         \hline 
    \end{tabular}
    \label{tab:challenge-stats}
\end{table}
\hl{During the analysis of these previous challenges, we already had several findings that were consistent with what we present in this paper. These include the winning approaches' generalization ability on unseen datasets. However it was not clear which components in the AutoML workflow contributed the most, which we will clarify in this work thanks to extensive ablation studies.}
In this work, we focus on the final AutoDL challenge with all domains combined together. Some of the principal questions we aimed at answering in this challenge ended up being answered, with the help of fact sheets that participants filled out, and some from the post-challenge experiment, as detailed further in the paper. The main highlights are now briefly summarized. 

First of all, {\bf were the tasks of the challenge of a difficulty adapted to push the state-of-the-art in Automated Deep Learning?} On one hand {\bf YES}, since (1) the top two ranking participants managed to pass all final tests without code failure and delivered solutions on {\bf new tasks} (trained and tested without human intervention), performing significantly better than the baseline methods, within the time/memory constraints, and (2) all teams used Deep Learning as part of their solutions. This confirms that Deep Learning is well adapted to the chosen domains (CV, NLP, speech). As  further evidence that we hit the right level of challenge duration and difficulty, 90\% of teams found the challenge duration sufficient and 50\% of teams found the time and computational resources sufficient. On the other hand {\bf NO}, since (1) all of the top-9 teams used a domain-dependent approach, treating each data modality separately (i.e. using hard-coded \textit{if..else} clauses and will probably fail on new unseen domains such as other sensor data); and (2) the time budget was too constraining to do any Neural Architecture Search; and (3) complex heterogeneous ensembles including non Deep Learning methods were used. 

Secondly, {\bf was the challenge successful in fostering progress in ``any-time learning''?} The learning curve examples in Figures \ref{fig:learning-curves-viktor} and \ref{subfig:learning-curves-carla} show that for most datasets, convergence was reached within 20 minutes \hl{(more experimental results presented in Section} \ref{subsec:more-time}). A fast increase in performance early on in the learning curve demonstrates that the participants made a serious effort to deliver solutions quickly, which is an enormous asset in many applications needing a quick turnover and for users having modest computational resources. 

Finally, from the research point of view, a burning question is {\bf whether progress was made in ``meta-learning''}, the art of learning from past tasks to perform better on new tasks? There is evidence that the solutions provided by the participants generalize well to new tasks, since they performed well in the final test phase. To attain these results, seven out of the nine top ranking teams reported that they used the provided ``public'' datasets for meta-learning purposes. In Section \ref{subsec:ablation-study} we used ablation studies to evaluate the importance of using meta-learning and in Section \ref{subsec:automl-generalization} we analyzed how well the solutions provided meta-generalize.

Thus, while we are still far from an ultimate AutoML solution that learns from scratch for ALL domains (in the spirit of \cite{real_automl-zero_2020}), we made great strides with this challenge towards democratizing Deep Learning by significantly reducing human effort. The intervention of practitioners is reduced to formatting data in a specified way; we provide code for that at \url{https://autodl.chalearn.org}, as well as the code of the winners.

\hl{The contributions of this work are:}
\begin{itemize}
    \item We provide a viable and working example system that evaluates AutoML and AutoDL solutions, using average rank, multiple-task execution and any-time learning metric;
    
    \item We provide an end-to-end toolkit~\footnote{\url{https://github.com/zhengying-liu/autodl-contrib}} for formatting data into the AutoDL format used in the challenges, \hl{also allowing new users to contribute new tasks to our repository and use our ``AutoDL self service'' (see below)};
    
    \item All winning solutions' code is open-sourced and \hl{we provide an ``AutoDL Self-Service''}\footnote{\url{https://competitions.codalab.org/competitions/27082}} that facilitates the application of the top-1 winning solution (\participant{DeepWisdom}) for making predictions;
    
    \item We provide a detailed description of the winning methods and \hl{fit them into a common AutoML workflow, which suggests a possible direction of future AutoML systems};
    
    \item We carry out extensive ablation studies on various components of the winning teams and show the importance of meta-learning, ensembling and efficient data loading;
    
    \item \hl{We explore the possibility of combining different approaches for a stronger approach and it turns out to be hard, which suggests the local optimality of the winning methods;}
    
    \item We study the impact of some design choices (such as the \hl{time budget} and the parameter $t_0$) and justify these choices.
\end{itemize}



The rest of this work is organized as follows. In Section~\ref{sec:challenge-design}, we give a brief overview of the challenge design (see \cite{liu_autocv_2019} for detailed introduction). Then, descriptions of winning methods are given in Section \ref{sec:winning-approaches} and in Appendix. Post-challenge analyses, including ablation study results, is presented in Section \ref{sec:experimental-results}. Lastly, we conclude the work in Section \ref{sec:conclusion}.


\section{Challenge design} \label{sec:challenge-design}


\subsection{Data}

\begin{figure}
    \centering
    \includegraphics[scale=0.5]{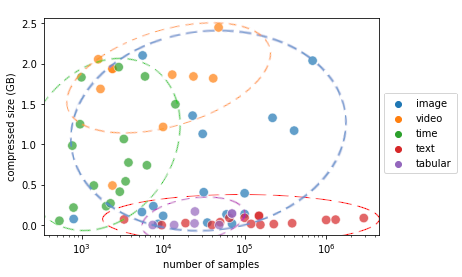}
    \caption{{\bf Distribution of AutoDL challenge dataset domains} with respect to compressed storage size in gigabytes and total number of examples for all 66 AutoDL datasets. We see that the text domain varies a lot in terms of number of examples but remains small in storage size. The image domain varies a lot in both directions. Video datasets are large in storage size in general, without surprise. Speech and time series datasets have fewer number of examples in general. Tabular datasets are concentrated and are small in storage size.}
    \label{fig:dataset-domains}
\end{figure}

In AutoDL challenges, {\bf raw data} (images, videos, audio, text, etc) are provided to participants formatted in a uniform tensor manner (namely TFRecords, a standard generic data format used by TensorFlow).
\footnote{\hl{To avoid privileging a particular type of Deep Learning framework, we also provided a data reader to convert the data to PyTorch format.}}
For images with native compression formats (e.g. JPEG, BMP, GIF), we directly use the bytes. Our data reader decodes them on-the-fly to obtain a 4D tensor. Video files in mp4/avi format (without the audio track) are used in a similar manner. For text datasets, each example (i.e. a document) is a sequence of integer indices. Each index corresponds to a word (for English) or character (for Chinese) in a vocabulary given in the metadata. For speech datasets, each example is represented by a sequence of floating numbers specifying the amplitude at each timestamp, similar to uncompressed WAV format. Lastly, tabular datasets' feature vector representation can be naturally considered as a special case of our 4D tensor representation. 

For practical reasons, each dataset was kept under 2.5 GB, which required sometimes reducing image resolution, cropping, and/or downsampling videos. We made sure to include application domains in which the scales varied a lot. We formatted around 100 datasets in total and used 66 of them for AutoDL challenges: 17 image, 10 video, 16 text, 16 speech and 7 tabular. The distribution of domain and size is visualized in Figure \ref{fig:dataset-domains}. All datasets marked public can be downloaded on corresponding challenge websites \footnote{\url{https://autodl.lri.fr/competitions/162}} and  information on some meta-features of all AutoDL datasets can be found on the 
``Benchmark'' page\footnote{\url{https://autodl.chalearn.org/benchmark}} of our website. All tasks are supervised multi-label classification problems, i.\,e. data samples are provided in pairs $\{X, Y\}$, $X$ being an input 4D tensor of shape (time, row, col, channel) and $Y$ a target binary vector (withheld from in test data). 
\hl{We have carefully selected the datasets out of 100 possibilities using two criteria: (1) having a high variance in the scores obtained by different baselines (modelling difficulty) and (2) having a relatively large number of test examples to ensure reasonable error bars (at least 1 significant digit)} \cite{guyon_what_1998}.



For the datasets of AutoDL challenge, we are not releasing their identities at this stage to allow us reusing them in future challenges. 
Some potential uses are discussed in Section \ref{sec:conclusion}. 
However, we summarize their name, domain and other meta-features in Table \ref{tab:all-datasets}. These datasets will appear in our analysis frequently.

\begin{table*}
\caption{{\bf Datasets of the AutoDL challenge,} for both phases.
The final phase datasets ({\bf meta-test} datasets) vary a lot in terms of number of classes, number of training examples, and tensor dimension, compared to those in the feedback phase. This was one of the difficulties of the AutoDL challenge. ``chnl'' codes for channel,  ``var'' for variable size, ``CE pair'' for ``cause-effect pair''. More information on all 66 datasets used in AutoDL challenges can be found at \url{https://autodl.chalearn.org/benchmark}. 
}
\label{tab:all-datasets}
\centering
\begin{tabular}{|c|c|c|c|c|c|c|c|c|c|c|c|}
\hline
     & & & & &Class &\multicolumn{2}{c|}{Sample number}&\multicolumn{4}{c|}{Tensor dimension} \\
    
     \# & Dataset & Phase & Topic  & Domain & num. & train & test & time & row & col & chnl  \\ \hline \hline
     
     \rowcolor{LightCyan}
     \rownumber &  Apollon & feedback & people  & image & 100 & 6077 & 1514 & 1 & var & var & 3 \\ \hline
     \rowcolor{LightCyan}
     \rownumber & Monica1 & feedback & action & video &20 & 10380 & 2565 & var & 168 & 168 & 3 \\ \hline
     \rowcolor{LightCyan}
     \rownumber & Sahak & feedback & speech & time & 100 & 3008 & 752 & var & 1 & 1 & 1 \\ \hline
     \rowcolor{LightCyan}
     \rownumber & Tanak & feedback & english & text & 2 & 42500 & 7501 & var & 1 & 1 & 1 \\ \hline
     \rowcolor{LightCyan}
     \rownumber & Barak  & feedback & CE pair & tabular & 4 & 21869 & 2430 & 1 & 1 & 270 & 1 \\ \hline
    
     \rowcolor{Cyan}
     \rownumber &  Ray & final & medical  & image & 7 & 4492 & 1114 & 1 & 976 & 976 & 3 \\  \hline
     \rowcolor{Cyan}
     \rownumber & Fiona & final & action & video &6 & 8038 & 1962 & var & var & var & 3  \\   \hline
     \rowcolor{Cyan}
     \rownumber & Oreal & final & speech & time & 3 & 2000 & 264 & var & 1 & 1 & 1 \\ \hline
     \rowcolor{Cyan}
     \rownumber & Tal & final & chinese & text & 15 & 250000 & 132688 & var & 1 & 1 & 1 \\ \hline
     \rowcolor{Cyan}
     \rownumber & Bilal  & final & audio & tabular & 20 & 10931 & 2733 & 1 & 1 & 400 & 1 \\ \hline
     
     \rowcolor{Cyan}
     \rownumber &  Cucumber & final & people  & image & 100 & 18366 & 4635 & 1 & var & var & 3 \\  \hline
     \rowcolor{Cyan}
     \rownumber & Yolo & final & action & video &1600 & 836 & 764 & var & var & var & 3  \\   \hline
     \rowcolor{Cyan}
     \rownumber & Marge & final & music & time & 88 & 9301 & 4859 & var & 1 & 1 & 1 \\ \hline
     \rowcolor{Cyan}
     \rownumber & Viktor & final & english & text & 4 & 2605324 & 289803 & var & 1 & 1 & 1 \\ \hline
     \rowcolor{Cyan}
     \rownumber & Carla  & final & neural & tabular & 2 & 60000 & 10000 & 1 & 1 & 535 & 1 \\ \hline

     \hline
\end{tabular}
\end{table*}

\subsection{Blind testing} 
\label{subsec:testing}
A hallmark of the AutoDL challenge series is that the code of the participants is blind tested, without any human intervention, in uniform conditions imposing restrictions on training and test time and memory resources, to push the state-of-the-art in automated machine learning. The challenge had 2 phases:
\begin{enumerate}
\item A {\bf feedback phase} during which methods were trained and tested on the platform on {\bf five practice datasets}\hl{, without any human intervention}. During the feedback phase, the participants could make several submissions per day and get immediate feedback on a leaderboard. The feedback phase lasted 4 months. Obviously, since they made so many submissions, the participants could to some extent get used to the feedback datasets. For that reason, we also had:
\item A {\bf final phase} using {\bf ten fresh datasets}. Only ONE FINAL CODE submission was allowed in that phase. 
\end{enumerate}
Since this was a complete blind evaluation during BOTH phases, we provided additional ``public'' datasets for practice purposes and to encourage meta-learning.

We ran the challenge on the CodaLab platform (\url{http://competitions.codalab.org}), which is an open source project of which we are community lead. CodaLab is free for use for all. We use to run the cALCulations a donation of Google of \$100,000 cloud credits. We prepared a docker including many machine learning toolkits and scientific programming utilities, such as Tensorflow, PyTorch and scikit-learn. We ran the jobs of the participants in virtual machines equipped with NVIDIA Tesla P100 GPUs. These virtual machines run CUDA 10 with drivers cuDNN 7.5 and 4 vCPUs, with 26 GB of memory, 100 GB disk. One VM was entirely dedicated to the job of one participant during its execution. Each execution must run in less than 20 minutes (1200 seconds) for each dataset.

\subsection{Metric} \label{subsec:metric}

\begin{figure}
    \centering
    \includegraphics[width=\linewidth]{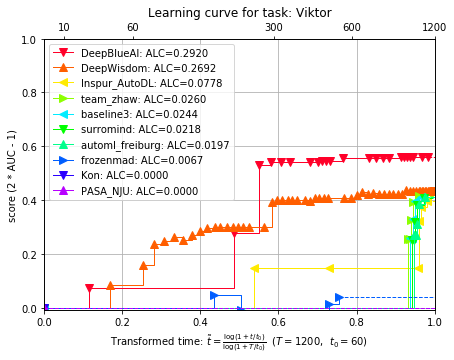}
    \caption{{\bf Learning curves of top-9 teams} (together with one baseline) on the text dataset \task{Viktor} from the AutoDL challenge final phase. We observe different patterns of learning curves, revealing various strategies adopted by participating teams. The curve of \participant{DeepWisdom} goes up quickly at the beginning but stabilizes at an inferior final performance (and also inferior any-time performance) than \participant{DeepBlueAI}. 
    In terms of number of predictions made during the whole train/predict process (20 minutes), many predictions are made by \participant{DeepWisdom} and \participant{DeepBlueAI} but (much) fewer are made by the other teams. Finally, although different patterns are found, some teams such as \participant{team\_zhaw}, \participant{surromind} and \participant{automl\_freiburg} show very similar patterns. This is because all teams adopted a domain-dependent approach and some teams simply used the code of Baseline 3 for certain domains (text in this case).
    }
    \label{fig:learning-curves-viktor}
\end{figure}


\hl{ The AutoDL challenge encouraged learning in a short time period  both by imposing a small time budget of 20 minutes per task and by using an ``any-time learning'' metric. Specifically, within the time budget, the algorithm could make several predictions (as many as they wanted), along the whole execution. This allowed us to use as performance score} the Area under the Learning Curve (ALC): 
\begin{equation}
\label{eq:ALC-definition}
    \begin{aligned}
    ALC &= \int_0^1 s(t) d\tilde{t}(t) \\
    &= \int_0^T s(t) \tilde{t}'(t) dt \\
    &= \frac{1}{\log (1 + T/t_0)} \int_0^T \frac{s(t)}{ t + t_0} dt \\
    \end{aligned} 
\end{equation}
where $s(t)$ is the performance score (we used the NAUC score introduced below) at timestamp $t$ and $\tilde{t}$ is the transformed time
\begin{equation}
\label{eq:time-transformation}
    \tilde{t}(t) = \frac{\log(1 + t / t_0)}{\log (1 + T / t_0)}.
\end{equation}
Here $T$ denotes the time budget in seconds (e.g. $T=1200$) and $t_0$ is a pre-defined time parameter, also in seconds (e.g. $t_0=60$).
Examples of learning curves can be found in Figure \ref{fig:learning-curves-viktor}). The participants can train in increments of a chosen duration (not necessarily fixed) to progressively improve {\em performance}, until the time limit is attained. Performance is measured by the NAUC or {\em Normalized} Area Under ROC Curve (AUC) 
$$NAUC = 2 \times AUC - 1$$
averaged over all classes. \hl{More details of the challenge protocol and evaluation workflow can be found in Appendix {\ref{sec:protocol}} and in Figure {\ref{fig:protocol}}.} Multi-class classification metrics are not being considered, i.\,e. each class is scored independently. Since several predictions can be made during the learning process, this allows us to plot learning curves, i.\,e. ``performance'' (on test set) as a function of time.  Then for each dataset, we compute the Area under Learning Curve (ALC). The time axis is log scaled (with time transformation in (\ref{eq:time-transformation})) to put more emphasis on the beginning of the curve.
This way, we encourage participants to develop techniques that improve performance rapidly at the beginning of the training process. This should be important to treat large redundant and/or imbalanced datasets and small datasets alike,
e.\,g. by treating effectively redundancy in large training datasets or using learning machines pre-trained on other data if training samples are scarce.
Finally, in each phase, an overall rank for the participants is obtained by averaging their ALC ranks obtained on each individual dataset. The average rank in the final phase is used to determine the winners. 
\michael{Je ne trouve pas ça très clair surtout la phrase "This implies notably that removing a dataset from the evaluation can't help an algorithm top-ranked on this dataset to win."}
\zhengying{Adrien: could you clarify? }
\hl{The use of the average rank allows us to fuse scores of different scales. Also, this ranking method satisfies some desired theoretical properties; it is in particular \textit{consistent} } \cite{consistency_average_rank}: \hl{whenever the datasets are divided (arbitrarily) into several parts and the average rankings of those parts garner the same ranking, then an average ranking of the entire set of datasets also garners that ranking. This implies notably that removing a dataset from the evaluation can't help an algorithm top-ranked on this dataset to win. Moreover, this empirical study} \cite{brazdil_average_rank} \hl{suggests that average rank is satisfying, compared to other ranking methods, in terms of rank correlation with unseen tasks. We are running similar experiments on AutoDL data and those results hold.}

\subsection{Baseline 3 of AutoDL challenge}
\label{subsec:baseline3}

\isabelle{Now that the details about winning solutions are in appendix, it is odd to have so much details about baseline 3 here. This should probably be shortered and details put in appendix.}
\zhengying{Baseline 3 is the baseline that is mostly used by the participants and all top-3 winners use it. Thus a description of it actually describes a huge part of  all the winning approach. So I tend to leave it here. We can decide on this later if there is no space left. }

As in previous challenges (e.g. AutoCV, AutoCV2, AutoNLP and AutoSpeech), we provided 3 baselines (Baseline 0, 1 and 2) for different levels of use: Baseline 0 is just constant predictions for debug purposes, Baseline 1 a linear model, and Baseline 2 a CNN (see \cite{liu_autocv_2019} for details). In the AutoDL challenge, we provided additionally a \textbf{Baseline 3} which combines the winning solutions of previous challenges \hl{(i.e. Baseline 3 first infers the domain/modality from the tensor shape and then apply the corresponding winning solution on this domain)}. And for benchmarking purposes, we ran Baseline 3 on all 66 datasets in all AutoDL challenges (public or not) and the results are shown in Figure \ref{fig:baseline3-all-datasets}. Many participants used Baseline 3 as a starting point to develop their own method. For this reason,  we describe in this section the components of Baseline 3 in some details.

\subsubsection{Vision domain: winning method of AutoCV/AutoCV2}

%
The wining solution of AutoCV1 and AutoCV2 Challenges~\cite{liu_autocv_2019}, i.e., \textit{kakaobrain}, is based on Fast AutoAugment~\cite{lim_fast_2019}, which is a modified version of the AutoAugment~\cite{cubuk_autoaugment:_2018} approach. Instead of relying on human expertise, AutoAugment~\cite{cubuk_autoaugment:_2018} formulates the search for the best augmentation policy as a discrete search problem and uses Reinforcement Learning to find the best policy. The search algorithm is implemented as a Recurrent Neural Network (RNN) controller, which samples an augmentation policy $S$, combining image processing operations, with their probabilities and magnitudes. $S$ is then used to train a child network to get a validation accuracy $R$, which is used to update the RNN controller by policy gradient methods. 

Despite a significant improvement in performance, AutoAugment requires thousands of GPU hours even with a reduced target dataset and small network. In contrast, Fast AutoAugment~\cite{lim_fast_2019} finds effective augmentation policies via a more efficient search strategy based on density matching between a pair of train datasets, and a policy exploration based on Bayesian optimization over stratified $k$-folds splits of the training dataset. The winning team (\participant{kakaobrain}) of AutoCV  implemented a light version of Fast AutoAugment, replacing the 5-folds by a single fold search and using a random search instead of Bayesian optimization. The backbone architecture used is ResNet-18 (i.e., ResNet~\cite{he_deep_2015} with 18 layers).

\subsubsection{Text domain: winning method of AutoNLP}


For the {\em text domain}, Baseline 3 uses \hl{the code from the 2nd place team upwind\_flys in} AutoNLP since we found that \participant{upwind\_flys}'s code was easier to adapt in the challenge setting and gave similar performance to that of 1st place (\participant{DeepBlueAI}).

The core of \participant{upwind\_flys}'s solution is a meta-controller dealing with multiple modules in the pipeline including model selection, data preparation and evaluation feedback. For the data preparation step, to compensate for class imbalance in the AutoNLP datasets, \participant{upwind\_flys} first cALCulates the data distribution of each class in the original data. Then, they randomly sample training and validation examples from each class in the training set, thus balancing the training and validation data by up- and down-sampling. Besides, \participant{upwind\_flys} prepares a model pool including fast lightweight models like LinearSVC~\cite{fan_liblinear_nodate}, and heavy but more accurate models like LSTM~\cite{hochreiter1997long} and BERT~\cite{devlin_bert:_2018}. They first use light models (such as linear SVC), but the meta-controller switches eventually to other models such as neural networks, with iterative training. 
If the AUC drops below a threshold or drops twice in a row, the model is switched, or the process is terminated and the best model ever trained is chosen, when the pool is exhausted.




\subsubsection{Speech domain: winning method of AutoSpeech}
Baseline 3 uses the approach of the 1st place winner of the AutoSpeech challenge: \participant{PASA\_NJU}.
Interestingly, PASA\_NJU, has developed one single approach for the two sequence types of data, i.e. speech and text.
As time management is key for optimizing any time performance, as measured by the metric derived from the ALC, the best teams have experimented with various data selection and progressive data loading approaches. Such decisions allowed them to create a trade-off between accelerating the first predictions while ensuring a good and stable final AUC.  
For instance PASA\_NJU truncated speech samples from 22.5s to 2.5s, and started with loading 50\% of the samples for the 3 first training loops, however preserving a similar balance of classes, loading the rest of the data from the 4th training loop.  
As for feature extraction, MFCC (Mel-Frequency Cepstral Coefficients)~\cite{davis1980comparison} and STFT (Short-Time Fourrier Transform)~\cite{sift} are used.
In terms of model selection and architectures, \participant{PASA\_NJU} progressively increases the complexity of their model, starting with simple models like LR (Logistic Regression), LightGBM at the beginning of the training, combined later with some light weight pre-trained CNN models like Thin-ResNet-34 (ResNet~\cite{he_deep_2015} but with smaller numbers of filters/channels/kernels) and VggVox~\cite{Nagrani18}, finally (bidirectional) LSTM~\cite{hochreiter1997long}, with attention mechanism. This strategy allows to make fast early predictions and progressively improves models performance over time to optimize the anytime performance metric.



\subsubsection{Tabular domain}
As there were no previous challenge for the tabular domain in AutoDL challenge series, the organizers implemented a simple multi-layer perceptron (MLP) baseline. Tabular datasets consist of both continuous values and categories. Categorical quantities are converted to normalized indices\hl{, i.e. by dividing indices (starting from $1$) by the total number of categories}. Tabular domains may have missing values (missing values are replaced by zero) as well. 
Therefore, to cope with missing data, we designed a denoising autoencoder (DAE)~\cite{vincent2008extracting} able to interpolate missing values from available data. The architecture consists of a batch normalization layer right after input data, a dropout, 4 fully connected (FC) layers, a skip connection from the first FC layer to the 3rd layer and an additional dropout after 2nd FC layer. Then we apply a MLP classifier with 5 FC layers. All FC layers have 256 nodes (expect the last layers of DAE and classifier) with ReLU activation and batch normalization. We keep the same architecture for all datasets in this domain. DAE loss is a L1 loss on non-missing data and classifier loss is a sigmoid cross entropy.




\begin{figure*}[t!]
    \centering
    \begin{subfigure}[t]{0.4\textwidth}
        \centering
        \includegraphics[width=\linewidth]{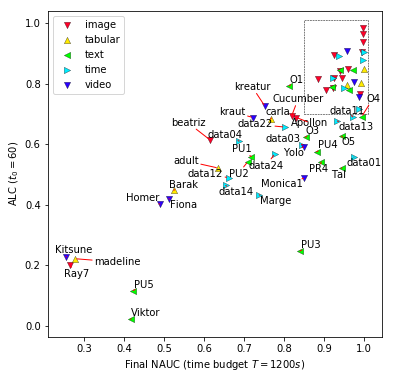} 
        \caption{{\bf Baseline 3} [top methods of previous challenges]}
        \label{subfig:baseline3-all-datasets}
    \end{subfigure}
    ~ 
    \begin{subfigure}[t]{0.4\textwidth}
        \centering
        \includegraphics[width=\linewidth]{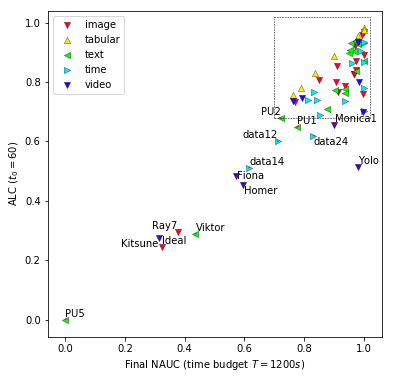} 

        \caption{{\bf DeepWisdom} \hl{ [winners final AutoDL challenge]}}
        \label{subfig:baseline3-all-datasets-zoomed}
    \end{subfigure}
    
    \caption{    \isabelle{Put the zoomed figure in appendix}
    \zhengying{Done}
    {\bf Performance gain in the Final AutoDL challenge} \hl{We plot ALC v.s. final NAUC performances of Baseline 3 and Deep Wisdom (the winners of the final AutoDL challenge) on ALL 66 datasets of the AutoDL challenge series benchmark. Different domains are shown with different markers. The dataset name is shown beside each point except the top-right area, which is zoomed in Appendix}  \ref{sec:zoomed-perf}\hl{, together with numerical values }(Table \ref{tab:baseline3-all-datasets}). \hl{DeepWisdom (AutoDL challenge winner) shows significant improvement over baseline 3, which included top methods of previous challenges in the series.}
    }
    \label{fig:baseline3-all-datasets}
\end{figure*}


\section{AutoDL challenge results}


The AutoDL challenge (the last challenge in the AutoDL challenges series 2019) lasted from 14 Dec 2019 (launched during NeurIPS 2019) to 3 Apr 2020. It has had a participation of 54 teams with 247 submissions in total and 2614 dataset-wise submissions. Among these teams, 19 of them managed to get a better performance (i.e. average rank over the 5 feedback phase datasets) than that of Baseline 3 in the feedback phase and entered the final phase of blind test. According to our challenge rules, only teams that provided a description of their approach (by filling out some fact sheets we sent out) were eligible for getting a ranking in the final phase. We received 8 copies of these fact sheets and thus only these 8 teams were ranked. These teams are (alphabetical order): \participant{DeepBlueAI}, \participant{DeepWisdom}, \participant{frozenmad}, \participant{Inspur\_AutoDL}, \participant{Kon}, \participant{PASA\_NJU}, \participant{surromind}, \participant{team\_zhaw}.
One team (\participant{automl\_freiburg}) made a late submission and isn't eligible for prizes but will be included in the post-analysis for scientific purpose.

\begin{figure*}[ht]
    \centering
    \includegraphics[width=\textwidth,
    ]{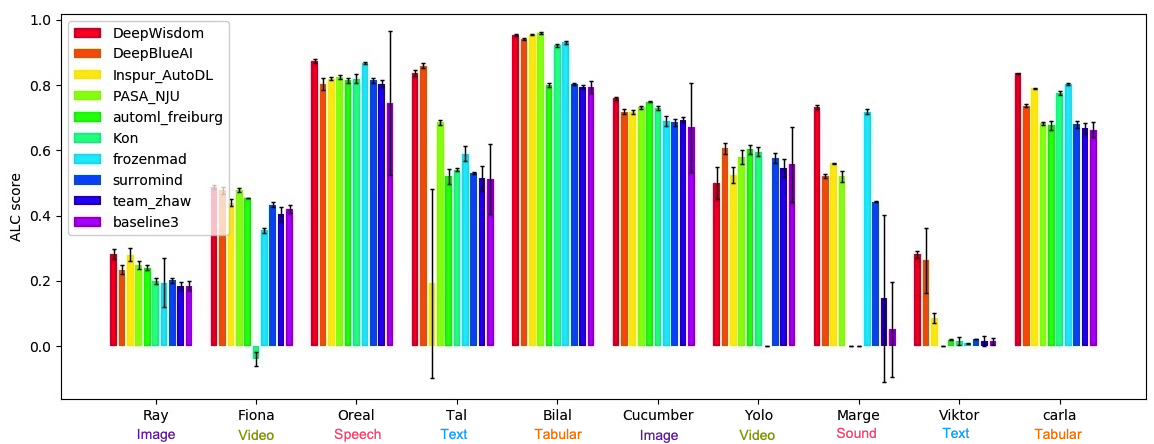}
    \caption{
    {\bf ALC scores of top 9 teams in AutoDL final phase} averaged over repeated evaluations (and Baseline 3, for comparison). The entry of top 6 teams are re-run 9 times and 3 times for other teams. Error bars are shown with (half) length corresponding to the standard deviation from these runs. Some (very rare) entries are excluded for computing these statistics due to failures caused by the challenge platform backend. The team ordering follows that of their average rank in the final phase. The domains of the 10 tasks are image, video, speech/times series, text, tabular (and then another cycle in this order). More information on the task can be found in Table \ref{tab:all-datasets}. 
    }
    \label{fig:top8-baseline3}
\end{figure*}

The final ranking is computed from the performances on the 10 unseen datasets in the final phase. To reduce the variance from diverse factors such as randomness in the submission code and randomness of the execution environment (which makes the exact ALC scores very hard to reproduce since the wall-time is hard to control exactly), we re-run every submission several times and average the ALC scores. The average ALC scores obtained by each team are shown in Figure \ref{fig:top8-baseline3} (the teams are ordered by their final ranking according to their average rank). From this figure, we see that some entries failed constantly on some datasets such as \participant{frozenmad} on \task{Yolo},
\participant{Kon} on \task{Marge} and
\participant{PASA\_NJU} on \task{Viktor}, due to issues in their code (e.g. bad prediction shape or out of memory error). In addition, some entries crashed only sometimes on certain datasets, such as \participant{Inspur\_AutoDL} on \task{Tal}, whose cause is related to some preprocessing procedure on text datasets concerning stop words. Otherwise, the error bars show that the performances of most runs are stable.

\isabelle{What does "statistically consistent" mean? I think this whole section could be improved. Justification of the ranking procedure and the stability of the ranking could come from Adrien's latest experiments.}

\zhengying{Changed ``statistically consistent'' to ``stable''. Adrien: could you integrate some of your results here?}



\section{Winning approaches}
\label{sec:winning-approaches}





A summary of the winning approaches on each domain can be found in Table \ref{tab:winning-solutions}. Another summary using a categorization by machine learning techniques can be found in Table \ref{tab:winning-solutions-by-technique}. We see in Table \ref{tab:winning-solutions} that almost all approaches used 5 different methods from 5 domains. For each domain, the winning teams' approaches are much inspired by Baseline 3 (see Section \ref{subsec:baseline3}).
\hl{For the two domains from computer vision (image and video), we spot popular backbone architectures such as ResNet}~{\cite{he_deep_2015}}
\hl{and its variants. Data augmentation techniques such as flipping, resizing are frequently used. Fast AutoAugment} {\cite{lim_fast_2019}}
\hl{from the AutoCV challenges winner solution is also popular. Pre-training (e.g. on ImageNet or Kinetics) is used a lot to accelerate training. For the speech domain and text domain, different feature extraction techniques using domain knowledge (such as MFCC, STFT, truncation) are used, as in the case of Baseline 3. For the tabular domain, more classical machine learning algorithms are used combined with intelligent data loading strategies.
}
In Table \ref{tab:winning-solutions-by-technique}, we see that almost all different machine learning techniques \hl{(such as meta-learning, preprocessing, HPO, transfer learning and ensembling)} are actively present and frequently used in all domains (exception some rare cases for example transfer learning on tabular data). 

\hl{By analyzing the workflow from all participating teams in final phase, we came up with an AutoML workflow shared by almost all teams (see Figure {\ref{fig:automl-workflow}}). We note that the module ``data'' includes not only traditional data of example-label pairs but also meta-data, metric, budgets and past performances. These are all potential useful information for meta-learning. Data are ingested by a Data Ingestor that consists of many sub-modules such as preprocessing, data augmentation, feature engineering and data loading management. Ingested data are then passed to the model/learner for learning and then they are both used by an Evaluator for evaluation (e.g. with a train/validation split). A Meta-learner can be applied (offline due to our challenge protocol) to accelerate all sub-modules of the model/learner AND optionally improved the Model Selector and the Data Ingestor, based on the meta-data of the current task and potentially a meta-dataset of prior tasks (e.g. those provided as public datasets). We believe that this AutoML workflow concisely summarizes the increasingly sophisticated AutoML systems found nowadays and provides the direction for a universal AutoML API design in the future (which is work in progress). This workflow will also be useful for the analysis in Section {\ref{sec:experimental-results}}.

The more detailed descriptions for the approaches of the top-3 winning teams and automl\_freiburg can be found in the Appendix.}



\begin{table*}[hbt]
    \setlength\tabcolsep{1.5pt}
    \scriptsize
    \caption{\textbf{Summary of the five top ranking solutions} and their average rank in the final phase. The participant's average rank (over all tasks) in the final phase is shown in parenthesis  (automl\_freibug and Baseline 3 were not ranked in the challenge). Each entry concerns the algorithm used for each domain and is of the form ``[preprocessing / data augmentation]-[transfer learning/meta-learning]-[model/architecture]-[optimizer]'' (when applicable).
    }
    \begin{tabular}{|C{2.0cm}|C{3cm}|C{3cm}|C{3.0cm}|C{3cm}|C{3cm}|}
    \hline
         Team  & image & video & speech & text & tabular  \\ \hline \hline
         1.DeepWisdom (1.8) & [ResNet-18 and ResNet-9 models] [pre-trained on ImageNet] & [MC3 model] [pre-trained on Kinetics] & [fewshot learning ] [LR, Thin ResNet34 models] [pre-trained on VoxCeleb2] & 
         [fewshot learning] [task difficulty and similarity evaluation for model selection] [SVM, TextCNN,[fewshot learning]  RCNN, GRU, GRU with Attention]& [LightGBM, Xgboost, Catboost, DNN models] [no pre-trained] \\ 
         \hline
         
         2.DeepBlueAI (3.5) & [data augmentation with Fast AutoAugment] [ResNet-18 model] & [subsampling keeping 1/6 frames] [Fusion of 2 best models ] & [iterative data loader (7, 28, 66, 90\%)] [MFCC and Mel Spectrogram preprocessing] [LR, CNN, CNN+GRU models] & [Samples truncation and meaningless words filtering] [Fasttext, TextCNN, BiGRU models] [Ensemble with restrictive linear model]& [3 lightGBM models] [Ensemble with Bagging] \\ 
         \hline
         
         3.Inspur\_AutoDL (4) & \multicolumn{4}{C{9cm}|}{Tuned version of Baseline 3} & [Incremental data loading and training][HyperOpt][LightGBM] \\ 
         \hline
         
         4.PASA\_NJU (4.1) & [shape standardization and image flip (data augmentation)][ResNet-18 and SeResnext50] & [shape standardization and image flip (data augmentation)][ResNet-18 and SeResnext50] & [data truncation(2.5s to 22.5s)][LSTM, VggVox ResNet with pre-trained weights of DeepWisdom(AutoSpeech2019) Thin-ResNet34] & [data truncation(300 to 1600 words)][TF-IDF and word embedding]& [iterative data loading] [Non Neural Nets models] [models complexity increasing over time] [Bayesian Optimization of hyperparameters] \\ 
         \hline
         
         5.frozenmad (5) & [images resized under 128x128] [progressive data loading increasing over time and epochs] [ResNet-18 model] [pre-trained on ImageNet] & [Successive frames difference as input of the model] [pre-trained ResNet-18 with RNN models] & [progressive data loading in 3 steps 0.01, 0.4, 0.7] [time length adjustment with repeating and clipping] [STFT and Mel Spectrogram preprocessing] [LR, LightGBM, VggVox models] & [TF-IDF and BERT tokenizers] [ SVM, RandomForest , CNN, tinyBERT ] & [progressive data loading] [no preprocessing] [Vanilla Decision Tree, RandomForest, Gradient Boosting models applied sequentially over time] \\
         \hline
         automl\_freiburg &  \multicolumn{2}{C{6cm}|}{Architecture and hyperparameters learned offline on meta-training tasks with BOHB. Transfer-learning on unseen meta-test tasks with AutoFolio. Models: EfficientNet [pre-trained on ImageNet with AdvProp], ResNet-18 [KakaoBrain weights], SVM, Random Forest, Logistic Regression} & \multicolumn{3}{C{9cm}|}{Baseline 3} \\
         \hline 
         Baseline 3 & [Data augmentation with Fast AutoAugment, adaptive input size][Pre-trained on ImageNet][ResNet-18(selected offline)]  & [Data augmentation with Fast AutoAugment, adaptive input size, sample first few frames, apply stem CNN to reduce to 3 channels][Pre-trained on ImageNet][ResNet-18(selected offline)]  & [MFCC/STFT feature][LR, LightGBM, Thin-ResNet-34, VggVox, LSTM]   & [resampling training examples][LinearSVC, LSTM, BERT]  & [interpolate missing value][MLP of four hidden layers] \\ 
         \hline
         
    \end{tabular}
\label{tab:winning-solutions}
\end{table*}

\begin{table*}[htb]
    \centering
    \setlength\tabcolsep{1.5pt}
    \scriptsize
    \caption{\textbf{Machine learning techniques} applied to each of the 5 domains considered in AutoDL challenge.
    }
    \begin{tabular}{|C{2.0cm}|C{3cm}|C{3cm}|C{3.0cm}|C{3cm}|C{3cm}|}
    \hline
         ML technique  & image & video & speech & text & tabular \\ \hline \hline

         Meta-learning & \multicolumn{5}{C{16cm}|} {
          Offline meta-training transferred with AutoFolio~\cite{lindauer_autofolio_2015} based on meta-features (\participant{automl\_freiburg}, for image and video) \linebreak
         Offline meta-training generating solution agents, searching for optimal sub-operators in predefined sub-spaces, based on dataset meta-data. (\participant{DeepWisdom})  \linebreak
         MAML-like method~\cite{finn_model-agnostic_2017} (\participant{team\_zhaw})} \\
         \hline
         
         Preprocessing & image cropping and data augmentation (\participant{PASANJU}), Fast AutoAugment (\participant{DeepBlueAI})
         & Sub-sampling keeping 1/6 frames and adaptive image size (\participant{DeepBlueAI})
         Adaptive image size 
         & MFCC, Mel Spectrogram, STFT &
         root features extractions with stemmer, meaningless words filtering (\participant{DeepBlueAI}) & 
         Numerical and Categorical data detection and encoding \\
         \hline
         
         Hyperparameter Optimization & \multicolumn{2}{C{6cm}|}{Offline with BOHB~\cite{falkner_bohb_nodate} (Bayesian Optimization and Multi-armed Bandit) (\participant{automl\_freiburg})
         Sequential Model-Based Optimization for General Algorithm Configuration (SMAC) \cite{smac-2017} (\participant{automl\_freiburg})} & Online model complexity adaptation (\participant{PASA\_NJU}) & Online model selection and early stopping using validation set (\participant{Baseline 3}(\participant{upwind\_flys})) &
         Bayesian Optimization (\participant{PASANJU}) \linebreak
         HyperOpt~\cite{bergstra_algorithms_2011} (\participant{Inspur\_AutoDL}) \\
         \hline
         
         Transfer learning & Pre-trained on ImageNet~\cite{russakovsky_imagenet_2015} (all teams except \participant{Kon})  & 
         Pre-trained on ImageNet~\cite{russakovsky_imagenet_2015} (all top-8 teams except \participant{Kon}) \linebreak 
         MC3 model pre-trained on Kinetics (\participant{DeepWisdom}) & ThinResnet34 pre-trained on VoxCeleb2 (\participant{DeepWisdom}) & BERT-like~\cite{devlin_bert:_2018} models pre-trained on FastText & (not applicable) \\
         \hline
         
         Ensemble learning & Adaptive Ensemble Learning (ensemble latest 2 to 5 predictions) (\participant{DeepBlueAI}) & Ensemble Selection~\cite{caruana_ensemble_2004} (top 5 validation predictions are fused) (\participant{DeepBlueAI}); Ensemble models sampling 3, 10, 12 frames (\participant{DeepBlueA})  &  
         last best predictions ensemble strategy (\participant{DeepWisdom}) \linebreak
         averaging 5 best overall and best of each model: LR, CNN, CNN+GRU (\participant{DeepBlueA}) & 
         Weighted Ensemble over 20 best models~\cite{caruana_ensemble_2004} (\participant{DeepWisdom}) & LightGBM ensemble with bagging method~\cite{ke_LightGBM_2017} (\participant{DeepBlueAI}), \linebreak
         Stacking and blending (\participant{DeepWisdom}) \\ 
         \hline

    \end{tabular}
\label{tab:winning-solutions-by-technique}
\end{table*}

\section{Post-challenge analyses}
\label{sec:experimental-results}

\begin{figure*}[ht]
    \centering
    \includegraphics[width=0.8\textwidth]{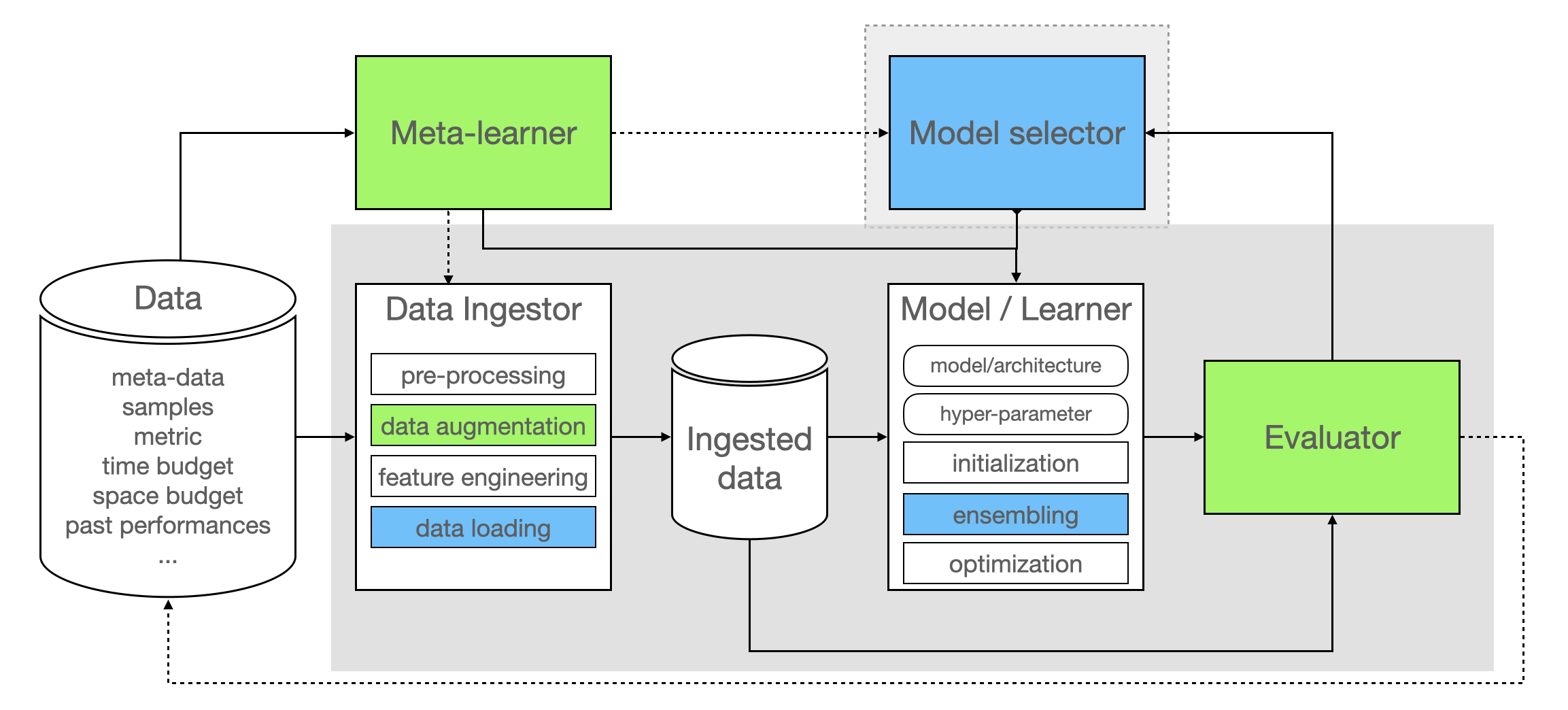}
    \caption{
    {\bf Global AutoML workflow shared by most participating teams.} \hl{Dotted arrows indicate optional (i.e. not used by everybody) connections between different components. Components in green are studied in ablation study in Section} \ref{subsec:ablation-study}. \hl{Components in blue are studied both in} Section \ref{subsec:ablation-study} and Section \ref{subsec:comb}. \hl{The modules in the grey shaded area are executed on the CodaLab competition platform }(i.e. \emph{online}).\hl{ Meta-learner runs in most cases offline (e.g. with the provided public datasets). Model selector can be executed online but pre-trained with meta-learner offline.}
    }
    \label{fig:automl-workflow}
\end{figure*}

\hl{
We carry out post-challenge analyses from different aspects to understand the results in depth and gain useful insights. One central question we ask ourselves is how the components (such as meta-learning, data loading and ensemble), in each approach, affect the final performance and whether one can combine these components from \emph{different} approaches and possibly obtain a stronger AutoML solution. These questions are addressed in Section}
\ref{subsec:ablation-study} and \ref{subsec:comb}. 
\hl{
For the reader to gain a global understanding of the relationship between different components, we visualize the overall AutoML workflow in Figure
}  \ref{fig:automl-workflow}.

\hl{
Apart from a local analysis of components, we also try to gain a global understanding of the AutoML generalization ability of all winning approaches in Section} \ref{subsec:automl-generalization}. 
\hl{The impact of some design choices of the challenge is studied in Section} \ref{subsec:more-time} and \ref{subsec:t0} and more discussions follow in later sections.

\subsection{Ablation study}
\label{subsec:ablation-study}

To analyze the contribution of different components in each winning team's solution, we asked 3 teams (\participant{DeepWisdom}, \participant{DeepBlueAI} and \participant{automl\_freiburg}) to carry out an ablation study, by removing or disabling certain component (e.g. meta-learning, data augmentation) of their approach. We will introduce in the following sections more details on these ablation studies by team and synthesize thereafter.

\subsubsection{\participant{DeepWisdom}} \label{subsubsec:deepwisdom}

According to the team \participant{DeepWisdom}, three of the most important components leading to the success of their approach are: meta-learning, data loading and data augmentation. For the ablation study, these components are removed or disabled in the following manner:


\begin{itemize}
    \item \textbf{Meta-learning (ML)}: Here meta-learning includes transfer learning, pre-train models, and hyperparameter setting and selection. Meta learning is crucial to both the final accuracy performance and the speed of train-predict lifecycle. For comparison we train models from scratch instead of loading pre-trained models for image, video and speech data, and use the default hyperparameter settings for text and tabular subtasks. 
    
    \item \textbf{Data Loading (DL)}: Data loading is a key factor in speeding up training procedures to achieve a higher ALC score. We improve data loading in several aspects. Firstly, we can accelerate decoding the raw data formatted in a uniform tensor manner to NumPy formats in a progressive way, and batching the dataset  for text and tabular data could make the conversion faster. Secondly, the cache mechanism is utilized in different levels of data and feature management, and thirdly, video frames are extracted in a progressive manner. 
    
    \item \textbf{Data Augmentation (DA)}: Fast auto augmentation, time augmentation and a stagewise spec\_len configuration for ThinResNet34~\cite{thinresnet} model are considered as data augmentation techniques for image, video and speech data respectively.
\end{itemize}

\begin{figure}
    \centering
    \includegraphics[width=\linewidth]{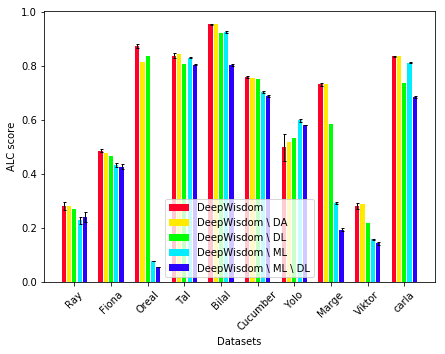}
    \caption{{\bf Ablation study for \participant{DeepWisdom}:} We compare different versions of \participant{DeepWisdom}'s approach, with one component of their workflow disabled.  ``DeepWisdom \textbackslash\, ML'' represents \participant{DeepWisdom}'s original approach but with Meta-Learning disabled. ``DA'' code for Data Augmentation and ``DL'' for Data Loading.  The method variants are ordered by their average rank from left to right. Thus we observe that removing Data Augmentation does not make a lot of difference, while removing both Meta-Learning and Data Loading impacts the solution a lot. See Section \ref{subsubsec:deepwisdom} for details.
    }
    \label{fig:deepwisdom}
\end{figure}


We carried out experiments on the 10 final phase datasets with the above components removed. The obtained ALC scores are presented in Figure \ref{fig:deepwisdom}. As it can be seen in  Figure~\ref{fig:deepwisdom}, Meta-Learning can be considered one of the most important single component in DeepWisdom's solution. Pre-trained models contribute significantly to both accelerating model training and obtaining higher AUC scores for image, video and speech data, and text and tabular subtasks benefit from hyperparameter setting such as model settings and learning rate strategies.
For image, we remove pre-trained models for both ResNet-18 and ResNet-9, which are trained on the ImageNet dataset with 70\% and 65\% top1 test accuracy; for video, we remove the parts of freezing and refreezing the first two layers. Then the number of the frames for ensemble models and replace MC3 model with ResNet-18 model. For speech, we do not load the pre-trained model which is pre-trained on VoxCeleb2 dataset, that is we train the ThinResNet34 model from scratch. 
For text, we use default setting, i.e. do not perform meta strategy for model selections and do not perform learning rate decay strategy selections. For tabular, with the experience of datasets inside and outside this competition, we found  two sets of parameters of LightGBM. 
The first hyperparameters focus on the speed of LightGBM training, it use smaller boost round and max depth, bigger learning rates and so on. While the second hyperparameters focus on the effect of LightGBM training, it can give us a generally better score. We use the default hyperparameters in LightGBM in the minus version.
    
Data Loading is a salient component for the ALC metric in any-time learning. For text, speech and tabular data, data loading speeds up NumPy data conversion to make the first several predictions as quickly as possible, achieving higher ALC scores. In the minus version, we convert all train TFRecord data to NumPy array in the first round, and ALC scores of nearly all datasets on all modalities decrease steadily compared with full version solution.

The data augmentation component also helps the ALC scores of several datasets. In the minus version for speech data we use the fixed spec\_len config, the default value is 200. Comparison on Marge and Oreal datasets is obvious, indicating that longer speech signal sequences could offer more useful information. Fast auto augmentation and test time augmentation enhance performance on image and video data marginally.

\subsubsection{\participant{DeepBlueAI}} \label{subsubsec:deepblueai}

According to the team \participant{DeepBlueAI}, three of the most important components leading to the success of their approach are: adaptive strategies, ensemble learning and scoring time reduction. For the ablation study, these components are removed or disabled in the following manner:


\begin{itemize}
    \item \textbf{Adaptive Strategies (AS)}: In this part, all adaptive parameter settings have been cancelled, such as the parameters settings according to the characteristics of datasets and the dynamic adjustments made during the training process. All relevant parameters are changed to default fixed values.
    \item \textbf{Ensemble Learning (EL)}: In this part, all the parts of ensemble learning are removed.
    Instead of fusing the results of multiple models, the model that performs best in the validation set is directly selected for testing.
    \item \textbf{Scoring Time Reduction (STR)}: In this part, all scoring time reduction settings were modified to default settings. Related parameters and data loading methods are same as those of baseline.
\end{itemize}

\begin{figure}
    \centering
    \includegraphics[width=\linewidth]{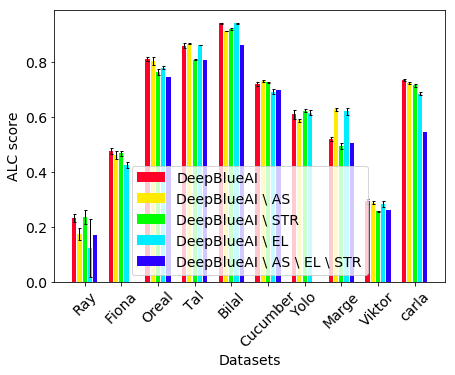}
    \caption{{\bf Ablation study for \participant{DeepBlueAI}:} Comparison of different versions of \participant{DeepBlueAI}'s approach after removing some of the method's components. ``DeepBlueAI \textbackslash\, AS'' represents their approach with Adaptive Strategy disabled. ``EL'' codes for Ensemble Learning and ``STR'' for Scoring Time Reduction. For each dataset, the methods are ordered by their average rank from left to right. While disabling each component separately yields moderate deterioration, disabling all of them yields a significant degradation in performance. See Section \ref{subsubsec:deepblueai}.}
    \label{fig:deepblueai}
\end{figure}


As it can be observed in Figure~\ref{fig:deepblueai},  
the results of DeepBlueAI have been greatly improved compared with those of DeepBlueAI \textbackslash AS \textbackslash EL \textbackslash STR (i.e., blue bar), indicating the effectiveness of the whole method.
After removing the AS, the score of most datasets has decreased, indicating that adaptive strategies are better than fixed parameters or models, and has good generalization performance on different datasets.
When STR is removed, the score of most datasets is reduced. Because the efficient data processing used can effectively reduce the scoring time, thereby improving the ALC score, which shows the effectiveness of the scoring time reduction.
After EL is removed, the score of the vast majority of datasets has decreased, indicating the effectiveness of ensemble learning to improve the results.

\subsubsection{\participant{automl\_freiburg}} 
\label{subsubsec:automl-freiburg}

According to the team \participant{automl\_freiburg}, two of the most important components leading to the success of their approach are: meta-learning and hyperparameter optimization. For the ablation study, these components are removed or disabled in the following manner:


\begin{itemize}
    \item \textbf{Meta-Learning with Random selector (MLR)}: This method randomly selects one configuration out of the set of most complementary configurations (Hammer, caltech\_birds2010, cifar10, eurosat).
    \item \textbf{Meta-Learning Generalist (MLG)}: This method does not use AutoFolio and always selects the generalist configuration that was optimized for the average improvement across all datasets.
    \item \textbf{Hyperparameter Optimization (HPO)}: Instead of optimizing the hyperparameters of the meta-selection model with AutoFolio, this method simply uses the default AutoFolio hyperparameters.
\end{itemize}

\begin{figure}
    \centering
    \includegraphics[width=\linewidth]{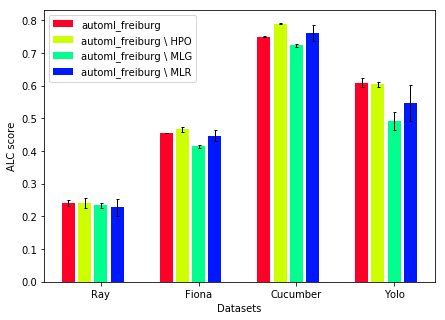}
    \caption{{\bf Ablation study for \participant{automl\_freiburg}:} Comparison of different versions of \participant{automl\_freiburg}'s approach. Since the approach addresses only computer vision tasks, only results on image datasets (\task{Ray}, \task{Cucumber}) and video datasets (\task{Fiona}, \task{Yolo}) are shown. Average and error bars of ALC scores are computed over 9 runs.  ``automl\_freiburg \textbackslash\, HPO'' represents \participant{automl\_freiburg}'s approach with default AutoFolio hyperparameters. Likewise, ``MLG'' stands for the generalist configuration and ``MLR'' for randomly selecting a configuration from the pool of the most complementary configurations. See Section \ref{subsubsec:automl-freiburg}.
    }
    \label{fig:automl-freiburg}
\end{figure}

As previously mentioned, \participant{automl\_freiburg} focused on the computer vision domain (i.e., datasets \task{Ray}, \task{Fiona}, \task{Cucumber}, and \task{Yolo}). The results of their ablation study, shown in Figure \ref{fig:automl-freiburg}, indicate that the hyperparameter search for the meta-model overfitted on the eight meta-train-datasets used (original vs HPO); eight datasets is generally regarded as insufficient in the realm of algorithm selection, but the team was limited by compute resources. However, the performance of the non-overfitted meta-model (HPO) clearly confirms the superiority of the approach over the random (MLR) and the generalist (MLG) baselines on all relevant datasets. More importantly, not only does this observation uncover further potential of \participant{automl\_freiburg}'s approach, it is also on par with the top two teams of the competition on these vision datasets: average rank 1.75 (\participant{automl\_freiburg}) versus 1.75 and 2.5 (\participant{DeepWisdom}, \participant{DeepBlueAI}). The authors emphasize that training the meta-learner on more than eight meta-train datasets could potentially lead to large improvements in generalization performance.
Despite the promising performance and outlook, results and conclusions should be interpreted conservatively due to the small number of meta-test datasets relevant to \participant{automl\_freiburg}'s approach.

\subsection{Combination study}
\label{subsec:comb}


In this section, instead of removing certain components for each winning method, we combine components from different teams. We start from a ``base'' method of one of the top ranking participants \participant{DeepWisdom} [\textbf{DW}], \participant{DeepBlueAI} [\textbf{DB]}, or \participant{automml\_freiburg} [\textbf{AF}], and we substitute (or add if absent) one or the key modules provided by another team. The design matrix is shown in Table \ref{tab:count-better-than-vanilla}. The lines represent the base solutions and the columns the models added or substitutes. Shaded matrix entries correspond to excluded cases: the modules considered were part of the base solution. This section is limited to the 6 image and video datasets of the AutoDL challenge for two reasons: (i) the [\textbf{AF}] team simply used baseline 3 for the other domains; (2) there were domain-specific architecture differences making difficult to conduct a more extensive study.

We focused on the following components, which demonstrated their effectiveness in our previous analyses, including the ablation study:
\begin{itemize}
    \item 
    The data loading (\textbf{DL}) component from \participant{DeepWisdom}, making good compromises between batch size, number of epochs, etc.;
    
    \item 
    The ensembling ({\textbf{EN}}) component from \participant{DeepBlueAI}. 
    an average of the last 5 predictions;
    \item 
    The meta-learning powered Hyper-Parameter Optimization (\textbf{HPO}) component from \participant{automml\_freiburg}, described in Sections \ref{subsubsec:automl-freiburg} and \ref{subsec:approach-automl-freiburg} with the usage of AutoFolio~\cite{lindauer_autofolio_2015}, using meta-features specific of the current task. The recommended set of hyperparameters is found by applying BOHB~\cite{falkner_bohb_nodate} off-platform on the public datasets.  
\end{itemize}

\begin{figure*}
    \centering
    \includegraphics[width=.9\linewidth]{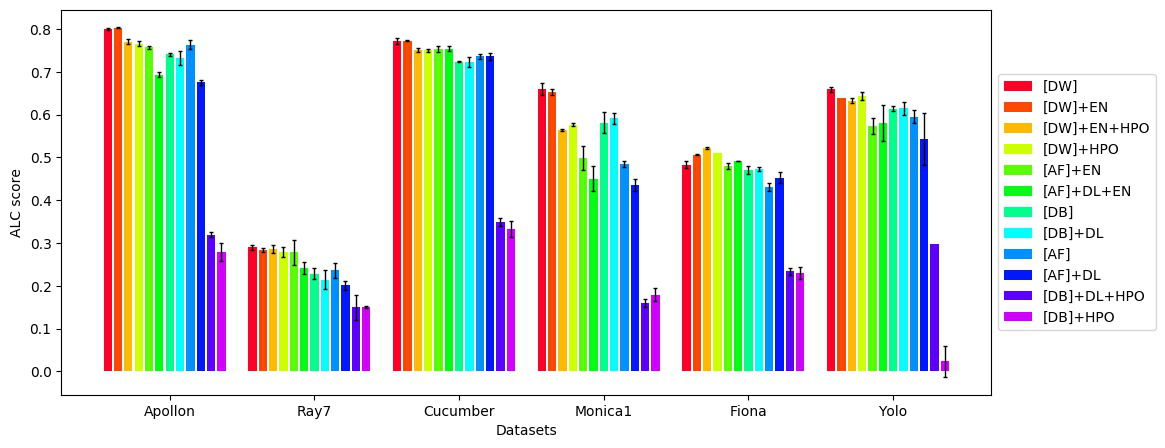}
    \caption{\textbf{\hl{Performances of different combinations of components}} from the three teams: \participant{DeepWisdom} (\textbf{DW}), \participant{DeepBlueAI} (\textbf{DB}) and  \participant{automl\_freiburg} (\textbf{AF}). \isabelle{C'est pas clair ce qu'est DL et ce qu'est DW} 
    The components we consider are (see Section \ref{subsec:comb}) Data Loading (\textbf{DL}) from \participant{DeepWisdom}, ensembling (\textbf{EN}) from \participant{DeepBlueAI} and hyperparameter optimization (\textbf{HPO}) from \participant{automl\_freiburg}. In the legend, [DW]+EN+HPO, for example, corresponds to the method of \participant{DeepWisdom} [DW] with the ensembling (EN) component replaced by that of \participant{DeepBlueAI}, and with the [HPO] component replaced by that of \participant{automl\_freiburg}.
    The methods are ordered by their average rank over all six considered tasks (3 image tasks and 3 video tasks), which are all from Table \ref{tab:all-datasets}. The error bars are computed from 3 repeated runs for each method. 
    \isabelle{C'est pas clair ce que ces combinaisons veulent dire. Qu'est-ce qui est utilisé pourles autres modules? ==> SOLUTION = mettre le tableau sur lequel nous avons travaill\'e aujourd'hui}
    We see that combining different components from different teams do not improve the ALC score in general.
    }
    \label{fig:comb}
\end{figure*}

We construct new combined methods using the following procedure: 
\begin{enumerate}
    \item Start from a base method, which is one of \participant{DeepWisdom} \textbf{[DW]}, \participant{DeepBlueAI} \textbf{[DB]} or \participant{automl\_freiburg} \textbf{[AF]};
    
    \item Replace one or two components in this base method by one of those from the other two teams, considering only the three components introduced above, i.e. \textbf{DL}, \textbf{EN} and \textbf{HPO}.
\end{enumerate}
\hl{For example, we use [AF]+DL+EN to denote the combined method that has}
\participant{automl\_freiburg}'s
\hl{ approach [AF] as base method, with the Data Loading (DL) component replaced by that of}
({\participant{DeepWisdom}}) 
\hl{and with the ensembling (EN) component replaced by that of}
\participant{DeepBlueAI}.
\hl{The question is how these components affect each other and whether one can construct a stronger method by combining different components. 
By plugging in one or two components from two other team, we manage to construct 3 new combinations for each team, making 9 new methods (and 12 methods in total with the three original approaches from each team). 
As} \participant{automl\_freiburg}\hl{'s approach focuses on image and video domains, we run the experiments on the 6 image and video datasets we used in the final AutoDL challenge. The results are presented in Figure}
\ref{fig:comb}.

\begin{table}[htb]
\centering
\begin{tabular}{|p{0.5cm}|p{0.4cm}|p{0.6cm}|p{0.6cm}|p{1cm}|p{1.3cm}|p{1.3cm}|}
\hline
         & +DL                      & +EN                      & +HPO                     & +DL+EN                                          & +DL+HPO                                         & +EN+HPO                                         \\ \hline
{[}DW{]} & \cellcolor[HTML]{959595} & 1                        & 1                        & \cellcolor[HTML]{333333}                        & \cellcolor[HTML]{333333}                        & 1                                               \\ \hline
{[}DB{]} & 0                        & \cellcolor[HTML]{959595} & 0                        & \cellcolor[HTML]{333333} & 0                                               & \cellcolor[HTML]{333333} \\ \hline
{[}AF{]} & 1                        & 4                        & \cellcolor[HTML]{959595} & 2                                               & \cellcolor[HTML]{333333} & \cellcolor[HTML]{333333} \\ \hline
\end{tabular}

\caption{
\textbf{Combination study design matrix.} \hl{We show the number of tasks (out of the 6 image/video tasks tested) on which the newly combined method obtains a better ALC score than that of the base method. Except for [AF], which is the weakest base method, little or no improvement is gained by borrowing modules from the other teams. The very simple (EN) module yields the most convincing improvement.}
}
\label{tab:count-better-than-vanilla}
\end{table}

\hl{From Figure} \ref{fig:comb}, 
\hl{
we see that combining different components to other teams harms the ALC performance in most cases.} \participant{DeepWisdom} (DW) is still ranked first (in terms of average rank over the 6 tasks) and performs better than those combined with \participant{DeepBlueAI}'s ensemble method (DW+EN) and with \participant{automl\_freiburg}'s HPO (DW+HPO) or both (DW+EN+HPO). 
We have similar observation for DB compared to DB+DL, DB+DL+HPO and DB+HPO. This indicates the integrity of each method and suggests that different components from one team are \emph{jointly} optimized and cannot be easily improved separately (i.e. locally optimal). 
An exception of this observation is the fact that AF+EN and AF+DL+EN perform better than AF. Actually, adding ensemble method generally improves the performance. 

Some other observations from Figure \ref{fig:comb} are:
\begin{itemize}
    \item Combining HPO to \participant{DeepBlueAI} (DB) significantly decreases the ALC. This can be seen from comparing DB+HPO (or DB+DL+HPO) to DB (or DB+DL). This means that applying AutoFolio from \participant{automl\_freiburg} doesn't necessarily improve ALC for any approach. We have consistent observations for \participant{DeepWisdom}, although with less radical impact;
    
    \item
    Applying data loading (DL) of \participant{DeepWisdom} to other teams do not improve the ALC in general, which is consistent with what we found in Figure \ref{fig:deepwisdom} on the image and video datasets (i.e. \task{Ray}, \task{Fiona}, \task{Cucumber} and \task{Yolo}). This means that for computer vision tasks, adjusting hyperparameters such as batch size and number of examples for preview only has limited effect on the ALC score. The potential gain in speed may be neutralized by its harm in accuracy;
    
    \item
    When applying two components from other teams, the changes are mostly consistent with the combined changes of adding one component one after another. For example, the performance of AF+DL+EN could be precisely inferred from the performance difference between AF+DL and AF and that between AF+EN and AF. This suggests that there may be approximately a locally linear dependence between the ALC performance and the considered components.
\end{itemize}

In summary, this limited set of combination experiments did not reveal a significant advantage of mixing and matching modules. The solution of the overall winner \participant{DeepWisdom} stands out.

\subsection{Varying the time budget}
\label{subsec:more-time}

\begin{figure}[htb!]
    \centering
    \includegraphics[width=0.8\linewidth]{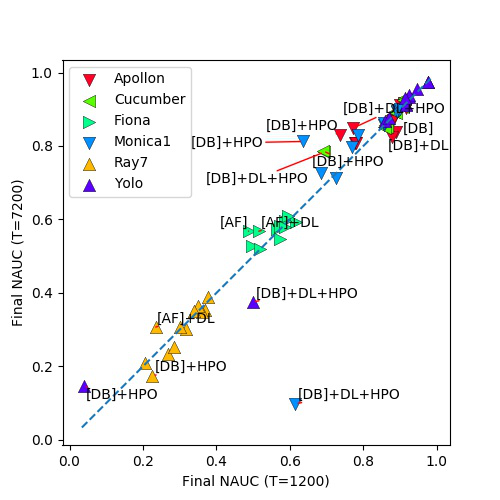}
    \caption{
    \textbf{\hl{Time budget comparison}. Comparison of final NAUC performance on }\hl{2h vs 20min}\hl{ time budget runs.}  Points with a NAUC difference (between two settings) larger than 0.05 are annotated. There are only 13 of these out of 72 points in total. 
    }
    \label{fig:more-time}
\end{figure}

\hl{Up till now, all our experiments are carried out within a time budget of 20 minutes, which may seem relatively small in this age of Big Data and Deep Learning. To investigate whether this time budget was sufficient and whether the approaches can perform better with a larger time budget, we run the same experiments as those in Section {\ref{subsec:comb}}
with exactly the same setting (the same algorithms and the same tasks) except that we change the time budget from 20 minutes ($T=1200$) to 2 hours ($T=7200$). And this time, we focus on the final NAUC instead of the ALC for a fair comparison. The results are visualized in Figure 
{\ref{fig:more-time}}.

In Figure {\ref{fig:more-time}}, each point corresponds to an approach-task pair such as (DB+HPO, Monica1). The tasks are shown in the legend and the approaches are annotated in some cases. We see that most points are close to the diagonal, which means that having a longer time budget does not improve the final NAUC performance in general. This suggests that most runs achieve convergence within 20 minutes, which is consistent with what we found when visualizing the learning curves (for example in Figure {\ref{fig:learning-curves-viktor}}). 
This finding further justifies our design choice of having a time budget of 20 minutes for all tasks.
Among all the 72 points, only 13 points out of them have a NAUC difference larger than 0.05 and these point are annotated with corresponding task names. Most of these annotated points correspond to the team DeepBlueAI combined with the HPO component from automl\_freiburg, meaning that this specific combination leads to a larger variance on the final NAUC. This can be explained by the fact that when AutoFolio (automl\_freiburg's HPO component which finds the prior task that is the most similar to the current one and recommends a hyperparameter configuration found offline for this prior task, see} Figure \ref{fig:asc} in Appendix \ref{subsec:approach-automl-freiburg}) \hl{chooses a hyperparameter configuration, the criterion it uses is based on the ALC performances obtained with automl\_freiburg's base method, which however is not what is being used since the base method is that of DeepBlueAI.
So AutoFolio is making performance predictions based on the wrong matrix of past performances} (details in Appendix \ref{subsec:approach-automl-freiburg}), \hl{corrupting the selection of hyperparameters and leading to a larger variance of the final NAUC score.
}
\isabelle{The last sentence does not make senses to me.}
\zhengying{Changed the last sentence. Is it clearer now?}
\isabelle{I still don't understand what AutoFolio does and why this contributes to the variance.}
\zhengying{Sorry the reference of the figure was wrong. And I added a sentence explaining what AutoFolio does.}
\michael{peut-être un meilleur terme que le vague "performance" pour parler du temps pour la dernière phrase sur AutoFolio car j'ai eu du mal à garder le lien avec le sujet de la sous-section en tête en première lecture. }

\subsection{AutoML generalization ability of winning methods}
\label{subsec:automl-generalization}

One crucial question for all AutoML methods is whether the method can have good performances on unseen datasets. If yes, we will say the method has \textit{AutoML generalization ability}. To quantitatively measure this ability, we propose to compare the average rank of all top-8 methods in both the feedback phase and the final phase, then compute the Pearson correlation (Pearson's $\rho$) of the 2 rank vectors (thus similar to Spearman's rank correlation~\cite{noauthor_spearmans_2020}). Concretely, let $r_X$ be the average rank vector of top teams in the feedback phase and $r_Y$ be that in the final phase, then the Pearson correlation is computed by $\rho_{X,Y} = \text{cov}(r_X, r_Y) / \sigma_{r_X} \sigma_{r_Y}$.

\begin{figure}[t]
    \centering
    \includegraphics[width=0.8\linewidth]{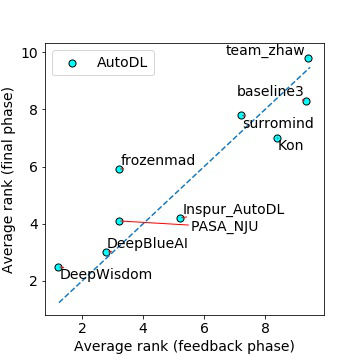}
    \caption{{\bf Task over-modeling:} We compare performance in the feedback and the final phase, in an effort to detect possible habituation to the feedback datasets due to multiple submissions. The average rank of the top-8 teams is shown. The figure suggests no strong over-modeling (over-fitting at the meta-learning level):  A team having a significantly better rank in the feedback phase than in the final phase would be over-modeling (far above the diagonal). The Pearson correlation is $\rho_{X,Y}=0.91$ and $p$-value $p=5.8 \times 10^{-4}$. }
    
    \label{fig:avg_rank-final-vs-feedback}
\end{figure}

The average ranks of top methods are shown in Figure \ref{fig:avg_rank-final-vs-feedback}, with a Pearson correlation $\rho_{X,Y}=0.91$ and $p$-value $p=5.8 \times 10^{-4}$. This means that the correlation is statistically significant and no leaderboard overfitting is observed. Thus the winning solutions can indeed generalize to \textit{unseen} datasets. Considering the diversity of the final phase datasets and the arguably out-of-distribution final-test meta-features shown in Table \ref{tab:all-datasets}, this is a feat from the AutoML community. Thus it's highly plausible that we are moving one step closer to a universal AutoML solution.

\subsection{Impact of $t_0$ in the ALC metric}
\label{subsec:t0}

\begin{figure*}[t!]
    \centering
    \begin{subfigure}[t]{0.3\textwidth}
        \centering
        \includegraphics[width=\linewidth]{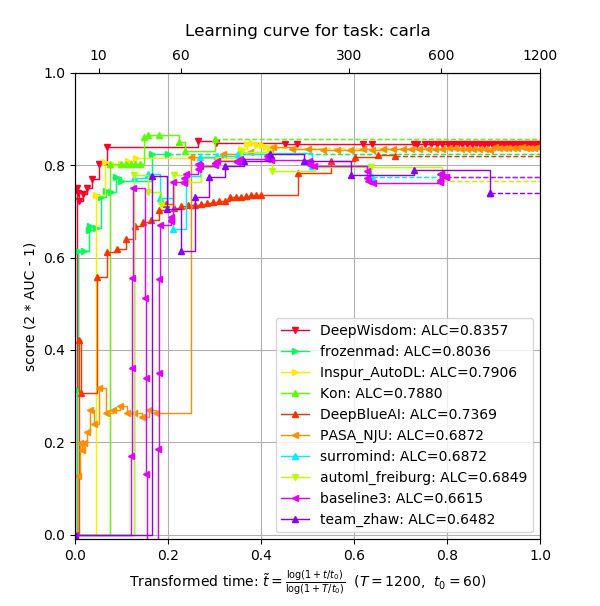}
        \caption{Learning curves for the task \task{Carla}}
        \label{subfig:learning-curves-carla}
    \end{subfigure}
    ~ 
    \begin{subfigure}[t]{0.3\textwidth}
        \centering
        \includegraphics[width=\linewidth]{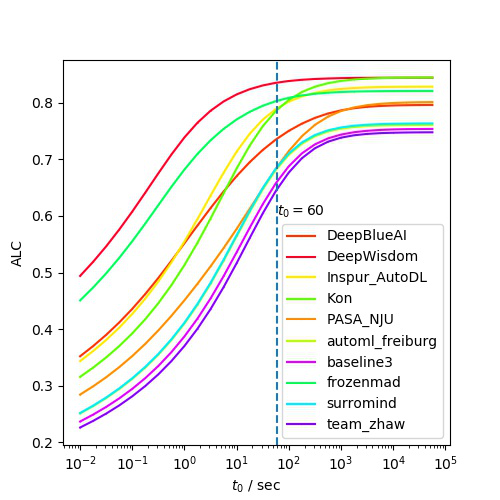}
        \caption{Impact of $t_0$ on the ALC scores for task \task{Carla}. }
        \label{subfig:t0-carla}
    \end{subfigure}
    ~
    \begin{subfigure}[t]{0.3\textwidth}
        \centering
        \includegraphics[width=\linewidth]{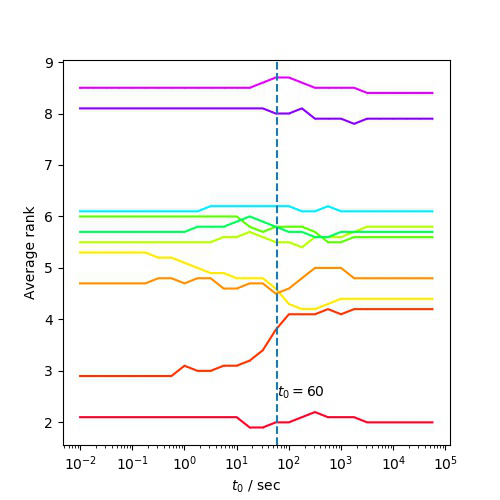}
        \caption{Average rank among AutoDL final phase participants, using different $t_0$. The legend is hidden and is the same as that of Figure \ref{subfig:t0-carla}.
        }
        \label{subfig:t0-avg-rank}
    \end{subfigure}
    
    \caption{{\bf Any-time learning {\em vs.} fixed-time learning:} We evaluate the impact of parameter $t_0$ on the ALC scores and the final rank. This parameter allows us to smoothly adjust the importance of the beginning of the learning curve (and therefore the pressure imposed towards achieving any-time learning). When t0 is small, the ALC puts more emphasis on performances at the beginning of the learning curve and thus favors fast algorithms.
When t0 is large, similar weight is applied on the whole learning curve, performances are uniformly averaged, so being a little bit slow at the beginning is not that bad, and it is more important to have good final performance when the time budget is exhausted (fixed-time learning).
 The tabular dataset \task{Carla} is taken as example. The fact that two learning curves cross each other is a necessary condition for the impact of $t_0$ on their ranking on this task.
 Learning curves of top teams on this dataset are shown in \ref{subfig:learning-curves-carla}. The impact of $t_0$ on the ALC scores of these curves is shown in \ref{subfig:t0-carla}. We see that when $t_0$ changes, the ranking among participants can indeed change, typically the ALC of \participant{frozenmad} is larger than that of \participant{Kon} but this is not true for large $t_0$. In \ref{subfig:t0-avg-rank}, the fact that the average rank (over all 10 final phase datasets) varies with $t_0$ also implies that $t_0$ can indeed affect the ranking of ALC on individual tasks. However, we see that the final ranking (i.e. that of average rank) is quite robust against changes of $t_0$. Very few exceptions exist such as \participant{PASA\_NJU} and \participant{Inspur\_AutoDL}.  Overall, $t_0$ proved to have little impact, particularly on the ranking of the winners, which is another evidence that top ranking participants addressed well the any-time learning problem.
    }
    \label{fig:impact-of-t0}
    
\end{figure*}

We recall that the Area under Learning Curve (ALC) is defined by

\begin{equation}
\label{eq:ALC}
    \begin{aligned}
    ALC &= \int_0^1 s(t) d\tilde{t}(t) \\
    &= \int_0^T s(t) \tilde{t}'(t) dt \\
    &= \frac{1}{\log (1 + T/t_0)} \int_0^T \frac{s(t)}{ t + t_0} dt \\
    \end{aligned} 
\end{equation}
where
\begin{equation}
\label{eq:time-transformation}
    \tilde{t}(t) = \frac{\log(1 + t / t_0)}{\log (1 + T / t_0)}
\end{equation}
Thus $t_0$ parameterizes a weight distribution on the learning curve for computing the ALC. When $t_0$ is small, the importance weight at the beginning of the curve is large. Actually when $t_0$ varies from 0 to infinity, we have 
$$\lim_{t_0 \to 0^+} ALC(t_0) = s(0)$$
and
$$\lim_{t_0 \to +\infty} ALC(t_0) = \frac{1}{T} \int_{0}^T s(t) dt.$$ So a different $t_0$ might lead to different ALC ranking even if the learning curve $s(t)$ is fixed. It is then to be answered whether the choice of $t_0=60$ in AutoDL challenge is reasonable. For this, we reflect the impact of $t_0$ on the ALC scores and the final average ranking in Figure \ref{fig:impact-of-t0}. Observation and discussion can be found in the caption. We conclude that $t_0$ does affect the ranking of ALC scores but the final ranking is robust to changes of $t_0$, justifying the choice of $t_0$ and the challenge setting.

\isabelle{Finalement j'ai supprimé la section de discussion: on n'a pas assez de choses interessantes à dire. J'ai tout mis dans la conclusion.}
\zhengying{D'accord}

\section{Conclusion}
\label{sec:conclusion}
\hl{
Automating Machine Learning and in particular Deep Learning, which has known recent successes in many application areas, is of central interest at the moment, to cut down the development cycle time, as well as to overcome the shortage of machine learning engineers. Our challenge series on AutoDL, and in particular the last one addressing the ubiquity of AutoDL solutions, allowed us to make great strides in this direction. To our knowledge, the solution of the winners, which was open-sourced, has no equivalent in academia or in the commercial arena. It is capable of training and testing effective models in 20 minutes to solve tensor-based multi-label classification problems. It has extensively been benchmarked on the 66 datasets of the entire challenge series, featuring a wide variety of types of data and dataset sizes. We have made the winners' solution available as a self-service}\footnote{\url{https://competitions.codalab.org/competitions/27082}}. \hl{Students using it in their projects have tested its efficacy on new tasks, demonstrating its ease-of-use. While this alone is a great outcome, our post-challenge analyses allowed us to pave the way to greater future improvements by analyzing module by module the contributions of the winning teams. First, it is remarkable that, in spite of the complexity of building a fully automated solution, and despite the fact that we did not impose any workflow or code skeleton, the top ranking teams converged towards a rather uniform modular architecture.  Our ablation studies revealed that the modules that may yield largest future improvement include ``meta-learning'' and ``ensembling'': Regarding} {\bf meta-learning},\hl{ at this stage, it is fair to say that strategies employed are effective, but not very sophisticated. They rely on pre-selecting off-platform, using provided ``public data'', one of the most promising neural architectures from the literature (typically based on ResNet for image, video, and speech, and BERT for text), pre-trained on large datasets (e.g. ImageNet for image and video). The submitted models were then fined-tuned on the challenge platform. One interesting twist has been the progressive tuning of weights starting from top layers, monitoring the depth of tuning as a hyperparameter. Most other hyperparameters however were frozen. There were pre-optimized outside the platform, which is another form of meta-learning. Our post-challenge studies did not reveal an improvement in performance when hyperparameters were optimized on the platform, using a state-of-the-art Bayesian optimization method. Regarding} {\bf ensembling}, \hl{ a wide variety of techniques were tried. Our ablation studies and combination studies revealed that one of the simplest methods is also the most effective: averaging predictions over the past few selected models. Second, our challenge put pressure on the participants to deliver fast solutions (in less than 20 minutes), and yielded technical advances in fast data loading, for instance. Our evaluation metric (Area under Learning curve) had two parameters allowing us to monitor both the total time budget and the dilation of the time axis (related to the importance put on getting good performance early on). The ranking of participants was robust against changes in both parameters and no significant improvements were gained by giving more time to the methods. On the flip side, 
the evaluation involving a learning curve as a function of time put emphasis on effectiveness of implementation, which were difficult to decouple from algorithm advances. In future challenges, we might want to factor out this aspect and are considering to rather use learning curves as a function of number of training examples or the computational operations (FLOPs), which should provide more reproducibility, more environment stability and less emphasis on engineering. Also, due to the small time budget of the AutoDL challenge, computationally expensive model search was not considered and could be the object of further work. To stimulate research in that direction, we have a Neural Architecture Search (NAS) challenge in preparation. To prevent participants to guess the data modality, the inputs are coded in a way, which makes it unobvious to recognize. This should avoid that participants leverage prior domain knowledge. Other challenges are under way. We started organizing a meta-learning challenge series }\footnote{https://metalearning.chalearn.org/} \hl{to evaluate meta-learning under controlled conditions rather than keeping it outside of the evaluation platform, as in the AutoDL challenge. Our goal is to encourage research on meta-learning in various settings, including few-shot learning. Beyond supervised learning, we are also interested in reinforcement learning. An AutoRL challenge is in preparation.}

\ifCLASSOPTIONcompsoc
  \section*{Acknowledgments}
\else
  \section*{Acknowledgment}
\fi

{\footnotesize
This work was sponsored with a grant from Google Research (Z\"urich) and additional funding from 4Paradigm, Amazon and Microsoft. 
It has  been partially supported by ICREA under the ICREA Academia programme. We also gratefully acknowledge the support of NVIDIA Corporation with the donation of the GPU used for this research.
The team \participant{automl\_freiburg} has partly been supported by the European Research Council (ERC) under the European Union’s Horizon 2020 research and innovation programme under grant no.\ 716721. Further, \participant{automl\_freiburg} acknowledges Robert Bosch GmbH for financial support. 
It received in kind support from the institutions of the co-authors. We are very indebted to Olivier Bousquet and Andr\'e Elisseeff at Google for their help with the design of the challenge and the countless hours that Andr\'e spent engineering the data format. The special version of the CodaLab platform we used was implemented by Tyler Thomas, with the help of Eric Carmichael, CK Collab, LLC, USA. Many people contributed time to help formatting datasets, prepare baseline results, and facilitate the logistics. We are very grateful in particular to: Stephane Ayache (AMU, France), Hubert Jacob Banville (INRIA, France), Mahsa Behzadi (Google, Switzerland), Kristin Bennett (RPI, New York, USA), Hugo Jair Escalante (IANOE, Mexico and ChaLearn, USA), Gavin Cawley (U. East Anglia, UK), Baiyu Chen (UC Berkeley, USA), Albert Clapes i Sintes (U. Barcelona, Spain), Bram van Ginneken (Radboud U. Nijmegen, The Netherlands), Alexandre Gramfort (U. Paris-Saclay; INRIA, France), Yi-Qi Hu (4paradigm, China), Tatiana Merkulova (Google, Switzerland), Shangeth Rajaa (BITS Pilani, India), Herilalaina Rakotoarison (U. Paris-Saclay, INRIA, France), Lukasz Romaszko (The University of Edinburgh, UK), Mehreen Saeed (FAST Nat. U. Lahore, Pakistan), Marc Schoenauer (U. Paris-Saclay, INRIA, France), Michele Sebag (U. Paris-Saclay; CNRS, France),
Danny Silver (Acadia University, Canada), Lisheng Sun (U. Paris-Saclay; UPSud, France), Wei-Wei Tu (4paradigm, China), Fengfu Li (4paradigm, China), Lichuan Xiang (4paradigm, China), Jun Wan (Chinese Academy of Sciences, China), Mengshuo Wang (4paradigm, China), Jingsong Wang (4paradigm, China), Ju Xu (4paradigm, China)}

\ifCLASSOPTIONcaptionsoff
  \newpage
\fi







\bibliographystyle{IEEEtran}
\bibliography{references}

%



%


\vskip 0pt plus -1fil

\begin{IEEEbiographynophoto}{Zhengying Liu} is a PhD student at Universit\'e Paris-Saclay, under the supervision of Isabelle Guyon. He received his bachelor degree at Peking University in fundamental mathematics and physics (double major) in 2013, master's degree in mathematics and computer science (double major) at Ecole polytechnique in 2017. His research interests lie in AutoML, deep learning and artificial intelligence in general including logic and automatic mathematical reasoning. He is one of the organizers of AutoDL challenges and has organized corresponding workshops at ECMLPKDD 2019 and NeurIPS 2019.
\end{IEEEbiographynophoto}

\vskip 0pt plus -1fil

\begin{IEEEbiographynophoto}{Adrien Pavao} is a PhD student at Universit\'e Paris-Saclay, under the supervision of Isabelle Guyon. He received his master's degree in computer science and machine learning at Universit\'e Paris-Saclay in 2019. His research topics include the methodology and experimental design in machine learning, the performance comparison between models and the organization of competitions.
\end{IEEEbiographynophoto}

\vskip 0pt plus -1fil

\begin{IEEEbiographynophoto}{Zhen Xu} is a research scientist at 4Paradigm, China. He works in Automated Machine Learning (AutoML) with applications in time series, graph, image, text, speech, etc. He also takes an active role in organizing AutoML/AutoDL challenges in top conferences, e.g. AutoCV, AutoSpeech, AutoSeries. Zhen received Engineering Degree (Diplôme d'Ingénieur) from Ecole polytechnique, Paris, majoring in Computer Science. He has a double diploma of Master in Applied Mathematics from University Paris-Sud, Orsay.
\end{IEEEbiographynophoto}

\vskip 0pt plus -1fil

\begin{IEEEbiographynophoto}{Sergio Escalera}
is Full Professor at Universitat de Barcelona and member of the Computer Vision Center at UAB. He leads the Human Behavior Analysis Group at UB and Computer Vision Center. He is series editor of The Springer Series on Challenges in Machine Learning. He is vice-president of ChaLearn Challenges in Machine Learning, leading ChaLearn Looking at People events. He is also member of the European Laboratory for Learning and Intelligent Systems. His research interests include automatic deep learning and analysis of humans from visual and multi-modal data, with special interest in inclusive, transparent, and fair affective computing and people characterization: personality and psychological profile computing.
\end{IEEEbiographynophoto}

\vskip 0pt plus -1fil

\begin{IEEEbiographynophoto}{Isabelle Guyon}
is Full Professor of Data Science and Machine Learning at Universit\'e Paris-Saclay, head master of the CS Artifical Intelligence master program, and researcher at INRIA.
She is also founder and president of ChaLearn, a non-profit dedicated to organizing challenges in Machine Learning and community lead on the development of the competition platform CodaLab.
She was co-program chair of NeurIPS 2016 and co-general chair of NeurIPS 2017, and now serving on the board of NeurIPS. She is an AMIA and an ELLIS fellow 
and action editor at JMLR, CiML springer series editor, and BBVA award recipient.
\end{IEEEbiographynophoto}

\vskip 0pt plus -1fil

\begin{IEEEbiographynophoto}{Sebastien Treguer}
is a research engineer at INRIA Saclay (TAU team), he also serve at the board of directors of Chalearn,  non-profit dedicated to organizing challenges in Machine Learning. He is a co-organizers of AutoML, AutoDL, MetaDL challenges. He is also a reviewer for ECML PKDD.
His research interests lies at the cross roads of machine learning and neuroscience.
\end{IEEEbiographynophoto}

\vskip 0pt plus -1fil

\begin{IEEEbiographynophoto}{Julio C. S. Jacques Junior}
is a postdoctoral researcher at the Computer Science, Multimedia and Telecommunications department at Universitat Oberta de Catalunya (UOC), within the Scene Understanding and Artificial Intelligence (SUNAI) group. He also collaborates within the Computer Vision Center (CVC) and Human Pose Recovery and Behavior Analysis (HUPBA) group at Universitat Autonoma de Barcelona (UAB) and University of Barcelona (UB), as well as within ChaLearn Looking at People.
\end{IEEEbiographynophoto}

\vskip 0pt plus -1fil

\begin{IEEEbiographynophoto}{Meysam Madadi}
obtained his MS degree and PhD in Computer Vision at the Universitat Autònoma de Barcelona (UAB) in 2013 and 2017, respectively. He is currently a postdoc researcher at Computer Vision Center (CVC), UAB. He has been a member of Human Pose Recovery and Behavior Analysis (HUPBA) group since 2012. His main interest is deep learning, computer vision, human pose estimation and garment modeling.
\end{IEEEbiographynophoto}

\vskip 0pt plus -1fil

\begin{IEEEbiographynophoto}{automl\_freiburg}
The automl\_freiburg team at the University of Freiburg (and nowadays also at the Leibniz University Hannover; Germany) was founded in 2015 and won several tracks of the first and second AutoML challenge. Members of the current challenge team are Fabio Ferreira, Danny Stoll, Arber Zela, Thomas Nierhoff, Prof. Marius Lindauer and Prof. Frank Hutter. Alumni of the challenge team include Matthias Feurer, Katharina Eggensperger, Aaron Klein and Stefan Falkner. Besides publications on AutoML at top journals and conferences, the group is well known for their open-source AutoML tools, such as Auto-Sklearn and Auto-PyTorch, see \url{www.automl.org}.
\end{IEEEbiographynophoto}

\vskip 0pt plus -1fil

\begin{IEEEbiographynophoto}{DeepBlueAI}
Team leader Zhipeng Luo received the M.S. degree from Peking University. He has nearly 6 years of machine learning experience. He has rich practical experience in computer vision, data mining and natural language processing. He has won championships in many top conference competitions, including CVPR, ICCV, KDD, NerulIPS, SIGIR, ACM MM , WSDM, CIKM, PAKDD, IEEE ISI. Members of the DeepBlueAI team are Chunguang Pan, Ge Li, Jin Wang and Kangning Niu.
\end{IEEEbiographynophoto}

\vskip 0pt plus -1fil

\begin{IEEEbiographynophoto}{Lenovo\_AILab} team comes from Lenovo Research, Members of the current challenge team are Peng Wang, Fuwang Zhao, Yuwei Shang, Xinyue Zheng, Bofan Liu. The main research fields include automatic deep learning, meta learning and distributed deep learning.
\end{IEEEbiographynophoto}

\vskip 0pt plus -1fil

\begin{IEEEbiographynophoto}{DeepWisdom} is a joint team of DeepWisdom and Xiamen University under the guidance of Prof. Rongrong~Ji and Chenglin Wu. Prof. Ji is currently a Professor and the Director of the Intelligent Multimedia Technology Laboratory, and the Dean Assistant with the School of Information Science and Engineering, Xiamen University, Xiamen, China, with over 100 papers published in international journals and conferences. Chenglin Wu is CEO of DeepWisdom. Other members of the team are Yang Zhang, Huixia Li, Sirui Hong and Youcheng Xiong. DeepWisdom is to build AI with AI, see \url{http://fuzhi.ai/}.
\end{IEEEbiographynophoto}




\clearpage

\appendices

\section{Challenge Protocol Details}
\label{sec:protocol}

\begin{figure*}[htb!]
    \centering
    \includegraphics[width=.9\linewidth]{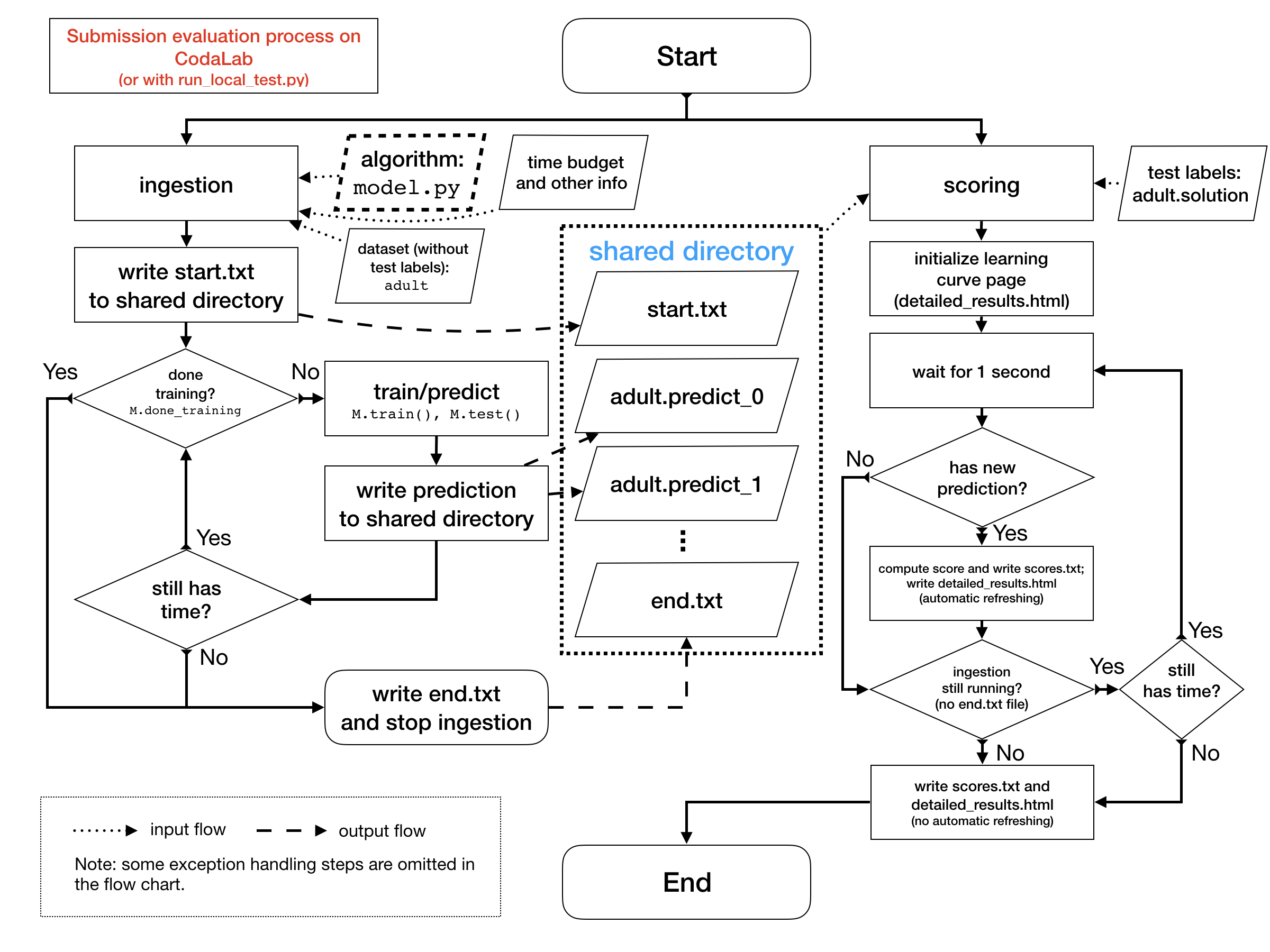}
    \caption{\textbf{\hl{The evaluation process of AutoDL challenge.}} The participant needs to prepare a ZIP file that contains at least a \texttt{model.py} file that implements a \texttt{Model} class with a \texttt{train()} method and a \texttt{test()} method. Two processes (ingestion and scoring) are started at the beginning in parallel. Ingestion process digests data and participant's submission. It calls participant's train/predict functions and write predictions to a shared directory. Scoring process listens to this directory and evaluates the predictions on the fly by comparing them to the hidden ground truth. When there is no more time or all the training process has been finished, an ending signal is written and both processes terminate. A learning curve is drawn according to the performances of the predictions made and an ALC score is computed for ranking. Finally, the ranks of the participant among all participants over all tasks are averaged and this average rank is  used for final ranking.
    }
    \label{fig:protocol}
\end{figure*}

\hl{
As introduced in Section {\ref{subsec:testing}}, the competition consists of two phases: a feedback phase and a final phase. Participants should submit an AutoML model code instead of predictions. 

Code submitted is trained and tested automatically, without any human intervention. Code submitted on "All datasets" is run on all five feedback or final datasets in parallel on separate compute workers, each with its own time budget. 

The identities of the datasets used for testing on the platform are concealed. The data are provided in a raw form (no feature extraction) to encourage researchers to use Deep Learning methods performing automatic feature learning, although this is NOT a requirement. All problems are multi-label classification problems. The tasks are constrained by the time budget (20 minutes/dataset). 

The submission evaluation process is shown in Figure} \ref{fig:protocol}. \hl{Participants' submissions are basically \texttt{model.py} but they could include other modules/files. Two processes (ingestion and scoring) are started at the beginning in parallel. Ingestion process digests data and participant's submission. It calls participant's train/predict functions and write predictions to a shared directory. Scoring process listens to this directory and evaluates on the fly. When there is no more time or all the training process has been finished, an ending signal is written and both processes terminate. A simplified pseudo-code is listed below.
}

\begin{algorithm}
\SetAlgoLined
\tcc{Initialize participants' model}
M = Model(metadata=dataset\_metadata)\;
    
\tcc{Initialize timer}
remaining\_time = overall\_time\_budget\;
start\_time = time()\;
\While{not M.done\_training and remaining\_time $> 0$}{
  M.train(training\_data, remaining\_time)\;
  Update remaining\_time\;
  results = M.test(test\_data, remaining\_time)\;
  Update remaining\_time\;
  \tcc{To be evaluated by scoring}
  save(results)\;
 }
 \caption{AutoDL challenge's evaluation protocol.}
\end{algorithm}






\section{Performance details of Baseline 3 and DeepWisdom}
\label{sec:zoomed-perf}

\begin{figure*}[hbt!]
    \centering
    \begin{subfigure}[t]{0.4\textwidth}
        \centering
        \includegraphics[width=\linewidth]{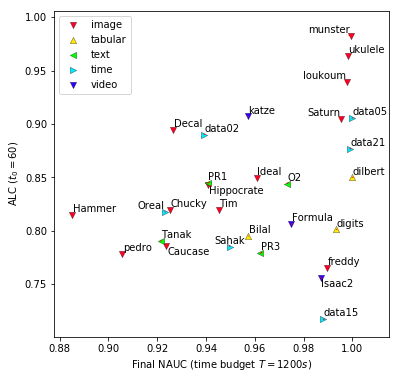} 
        \caption{{\bf Baseline 3}}
        \label{subfig:baseline3-all-datasets}
    \end{subfigure}
    ~ 
    \begin{subfigure}[t]{0.4\textwidth}
        \centering
        \includegraphics[width=\linewidth]{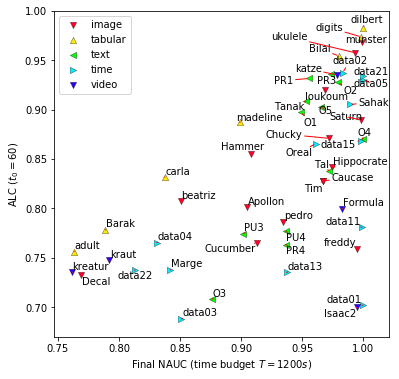} 

        \caption{{\bf DeepWisdom}}
        \label{subfig:baseline3-all-datasets-zoomed}
    \end{subfigure}
    
    \caption{{\bf Zoomed rectangular area of Figure} \ref{fig:baseline3-all-datasets}
    }
    \label{fig:baseline3-deepwisdom-all-datasets-zoomed}
\end{figure*}

To provide better visualization of the performances of Baseline 3 and DeepWisdom over the 66 AutoDL datasets, we show a zoomed version of the rectangular areas in Figure \ref{fig:baseline3-all-datasets} in Figure \ref{fig:baseline3-deepwisdom-all-datasets-zoomed}. 

\section{Description of winning methods}

We present in detail 
the winning solutions from top-3 winning teams (\participant{DeepWisdom}, \participant{DeepBlueAI} and \participant{PASA\_NJU}) and the team \participant{automl\_freiburg} which made a late submission in the feedback phase but ranked 5th in the final phase. We considered interesting to introduce \participant{automl\_freiburg}'s approach due to their contributions and for scientific purpose. 

\subsection{Approach of \participant{DeepWisdom} (1st prize)}

The team \participant{DeepWisdom} proposed a unified learning framework following a meta-learning paradigm. The framework consists of two parts: meta-train and meta-inference. The meta-train module takes as input the "public" datasets, which are augmented by the internal data augmentation engine, and the objective function (the ALC metric in the case of the challenge). The meta-trainer generates {\em solution agents}, whose objective is to search for best models, using search operators.
In the meta-inference step, a new task is processed taking in one dataset of the challenge. Initial metadata and seed data (few-shot samples) are acquired from the raw dataset. This constitutes the input of the {\em solution agents} obtained by meta-training. Solution workflow starts after taking in the seed input data, then it receives more raw data in a streaming way, and interacts with a whole set of tables for storage to cache intermediate results and models. Next, we explain the domain-specific contributions of \participant{DeepWisdom}.

In the image domain, ResNet-18 is used in the early stages of the training and then if the aspect ratio of images is between 0.5 and 2, it switched to ResNet-9 after 2 epochs because of the instability of ResNet-18 in later stages of training. When switching from ResNet-18 to ResNet-9, to reduce I/O cost, it doesn't share lower layers, but it caches the mini batches, which have been used for ResNet-18 training in GPU and reuse them for the initial training phase of ResNet-9, until all these mini batches are exhausted. The networks are fine-tuned by initialing from Imagenet pre-trained networks. However, for a fast transfer learning batch normalization and bias variables are initialized from scratch. To avoid overfitting, fast auto augmentation is used in the later training phase, which can automatically search for the best augmentation strategy, according to the validation AUC. As the searching process is quite time-consuming, it was done offline and then the resulted augmentation strategy was effectively applied during online training to increase the top-AUC.

In the video domain, a mixed convolution (MC3) network \cite{tran2018closer} is adopted which consists of 3D convolutions in the early layers and 2D convolutions in the top layers of the network. The network is pre-trained on the Kinetics dataset and accelerated transferring to other datasets by re-initializing linear weights and bias and freezing the first two layers. Due to the slower speed of 3D than 2D convolution, 3 frames are extracted at the early phase. Then for longer videos, an ensemble strategy is applied to combine best predictions from MC3 with 3-,10- and 12-frames data.

In the speech domain, a model search is applied in the meta-training part and LogisticRegression and ThinResnet34 \cite{thinresnet} achieve best performance in non-neural and neural models, respectively. The meta-trainer firstly learned that validating in the beginning was wasting the time budget without any effect on ALC, thus the evaluation agent did not validate when model was fitting new streaming data. Secondly, if amount of training samples was not very large, evaluation metric on training data could avoid overfitting partly while last best predictions ensemble strategy was applied.

In the text domain, they decode maximum 5500 samples for each round. Various data preprocessing methods are applied, adapted to the data structure of the public data provided in the first development phase, including email data structure preprocessing, word frequency filtering and word segmentation. After tokenization and sequence padding, both pre-trained and randomly initialized word embedding (with various dimensions) are used as word features. The meta-trainer includes several solutions such as TextCNN, RCNN, GRU, and GRU with attention \cite{kim2014convolutional}, \cite{rcnn}. Hyperparameters are set after a neural network architecture is selected. Also a weighted ensembling is adopted among top 20 models based AUC scores.

Finally, in the tabular domain, they batch the dataset and convert tfdatasets to NumPy format progressively, a weighted ensembling is applied based on several optimized models including LightGBM, Catboost, Xgboost and DNN on the offline datasets. To do so, data is split to several folds. Each fold has a training set and two validation sets. One validation set is used to optimize model hyperparameters and other set to compute ensembling weights.

\subsection{Approach of \participant{DeepBlueAI} (2nd prize)}


The DeepBlueAI solution is a combination of methods that are specific to each modality. Nevertheless, three concepts are applied across all modalities:
1) optimizing time budget by reducing the time for data processing, start with light models and parameters setting to accelerate first predictions;
2) dataset adaptive strategies and 3) ensemble learning.

For images, the DeepBlueAI team applies a strategy adapted to each specific dataset. They apply a pre-trained ResNet-18 model. The dataset adaptive strategy is not applied to model selection but to parameters settings including: image size, steps per epoch, epoch after which starting validating and fusing results. With the aim to optimize for final AUC, and make results more stable, they apply a progressive ensemble learning method, i.e. for epochs between 5 to 10, the latest 2 predictions are averaged, while after 10 epochs the 5 latest predictions are averaged. When the score on validation set improves a little,  a data augmentation strategy is adopted by searching for the most suitable data augmentation strategy for each image dataset with a small scale version of Fast AutoAugment \cite{lim2019fast} limiting the search among 20 iterations in order to preserve more time for training.

For video, ResNet-18 is used for classification. In the search for a good trade-off between cALCulation speed and classification accuracy, 1/6 of the frames with respect to the total number are selected. For datasets with a large number of categories, image size is increased to 128 to get more details out of it. During training, when the score of the validation set increases, predictions are made on the test set, and submitted as the average of the current highest 5 test results.
    

For speech, features are extracted with Mel spectrogram \cite{bridle1974experimental} for Logistic Regression (LR) model and MFCC \cite{davis1980comparison} for deep learning models. In order to accelerate the extraction long sequences are truncated but covering at least 90\% of the sequence. Then, to accelerate first score computation, training data are loaded progressively, 7\% for the first iteration, then 28\%, 66\% and then all data at 4th iteration, with care to balance multiple categories, to ensure the models can learn accurately. As for the models, LR is used for the first 3 iterations, then from the 4th iteration using all the data deep learning models, CNN and CNN+GRU \cite{cho2014learning} are employed. At the end, the overall 5 best models and the best version of each of the 3 models are averaged to build a final ensemble. The iterative data loading is especially effective on large dataset and plays a significant role in the performance measured by the metric derived from the ALC.
    
For text, the dataset size, text length and other characteristics are automatically obtained, and then a preprocessing method suitable for the dataset is adopted.
Long texts, over 6000 words are truncated, and NLTK stemmer is used to extract root features and filter meaningless words with frequency below 3.
As for model selection, FastText \cite{joulin2016bag}, TextCNN \cite{kim2014convolutional}, BiGRU \cite{cho2014learning} are used by their system that generate different model structures and set of parameters adapted to each dataset. The size of the dataset, the number of categories, the length of the text, and whether the categories are balanced are considered to generate the most suitable models and parameter settings.
    

For tabular, three directions are optimized: accelerating scoring time, adaptive parameter setting, ensemble learning.
    
Data is first split into many batches to significantly accelerate the data loading and converted from TFrecords to NumPy format. In terms of models, decision trees LightGBM are adopted to get faster scoring than with deep learning models. Because LightGBM supports continuous training, and the model learns faster in the early stage. During the training phase, earnings from the previous epochs are much higher than those from the latter. Therefore, a complete training is intelligently divided into multiple parts. The result is submitted after each part to obtain a score faster.

    In terms of adaptive parameter setting, some parameters are automatically set according to the size of data and the number of features of the tables. If the number of samples is relatively large, the ensemble fraction is reduced. If the original features of the sample are relatively large, the feature fraction is reduced. 
    A learning rate decay is applied, starting with a large value to ensure a speed up in the early training. 
    An automatic test frequency is adopted. Specifically, the frequency of testing is controlled based on training speed and testing speed. If the training is slow and the prediction is fast, the frequency of the test is increased. On the contrary, if training is fast and prediction is slow, the frequency is reduced. This strategy can improve to higher early scores.

    In order to improve generalization, multiple lightGBM models are used to make an ensemble with a bagging method.

\subsection{Approach of \participant{PASA\_NJU} (3rd prize)}
The \participant{PASA\_NJU} team modeled the problem as three different tasks:  CV (image and video), Sequence (speech and text) and Tabular (tabular domain). 

For the CV task, they preprocessed the data by analysing few sample instances of each dataset at training stage (such as image size, number of classes, video length, etc) in order to standardize the input shape of their model. Then, simple transformations (image flip) were used to augment the data. Random frames were obtained from video files and treated as image database. For both Image and Video tasks, ResNet-18 \cite{he_deep_2015} is used. However, SeResnext50 \cite{Hu_2018_CVPR} was used at later stages. Basically, they monitor the accuracy obtained by the ResNet-18 model and change the model to the SeResnext50 if no significant improvement is observed.

Speech and Text data are treated similarly, i.e., as a Sequence task. In a preprocessing stage, data samples are cut to have the same shape. Their strategy was to increase the data length as time passes. For example, they use raw data from 2.5s to 22.5s in speech task, and from 300 to 1600 words when Text data is considered. In both cases, hand-crafted feature extraction methods are employed. For speech data, mel spectrogram, MFCC and STFT \cite{sift} is used. When Text is considered, TF-IDF and word embedding is used. To model the problem, they employed Logistic Regression at the first stages and use more advanced Neural Networks at later stages, such as LSTM and Vggvox Resnet \cite{Nagrani18} (for speech data), without any hyperparameter optimization method. In the case of Vggvox Resnet, pre-trained model from \textit{Deepwisdom}'s team from AutoSpeech Challenge 2019~\cite{liu_autodl_2019} was used.

For Tabular data, they divided the entire process into three stages based on the given time budget, named Retrieve, Feature, and Model, and employed different models and data preprocessing methods at each stage, aiming to have quick responses at early stages. The main task of the Retrieve stage is to get the data and predict as soon as possible. Each time a certain amount of data is acquired, a model is trained using all the acquired
data. Thus, the complexity of the model is designed to increase with time. The main task of the Feature stage is to search for good features. As the Neural Feature Seacher(NFS) \cite{chen2019neural} method uses RNN as the controller to generate the feature sequence, they
used the same method and speed up the process by parallelizing it. Finally, at the Model stage, the goal is to search for a good model and hyperparameters. For this, they use hyperopt \cite{hyperopt}, which is an open-source package that uses Bayesian optimization to guide the search of hyperparameters.

\subsection{Approach of \participant{automl\_freiburg}}
\label{subsec:approach-automl-freiburg}

In contrast to other teams, \participant{automl\_freiburg} adopts a domain-independent approach but focused only the computer vision tasks (i.e. image and video datasets) of this challenge. While for all other tasks \participant{automl\_freiburg} simply submitted the baseline to obtain the baseline results, they achieved significant improvement on the computer vision tasks w.r.t. the baseline method. To improve both efficiency and flexibility of the approach, they first exposed relevant hyperparameters of the previous AutoCV/AutoCV2 winner code \cite{kakaobrain} and identified well-performing hyperparameter configurations on various datasets through hyperparameter optimization with BOHB~\cite{falkner_bohb_nodate}. They then trained a cost-sensitive meta-model~\cite{xu_satzilla2012_2012}  with AutoFolio \cite{lindauer_autofolio_2015} -- performing hyperparameter optimization for the meta-learner -- that allows to automatically and efficiently select a hyperparameter configuration for a given task based on dataset meta-features. The proposed approach on the CV task is detailed next.

First, they exposed important hyperparameters of the AutoCV/AutoCV2 winner's code \cite{kakaobrain} such as the learning rate, weight decay or batch sizes. Additionally, they exposed hyperparameters for the online execution (which were hard-coded in previous winner solution) that control, for example, when to evaluate during the submission and the number of samples used. To further increase the potential of the existing solution, they extended the configuration space to also include:
\begin{itemize}
    \item An EfficientNet~\cite{tan_efficientnet_2019} (in addition to  \participant{kakaobrain}'s\cite{kakaobrain} ResNet-18) pre-trained on ImageNet~\cite{russakovsky_imagenet_2015};
    \item The proportion of weights frozen when fine-tuning;
    \item Additional stochastic optimizers (Adam \cite{kingma_adam:_2014}, AdamW \cite{loshchilov-iclr19a},  Nesterov accelerated gradient \cite{nesterov83}) and learning rate schedules (plateau, cosine \cite{loshchilov-iclr17a});
    \item A simple classifier (either a SVM, random forest or logistic regression) that can be trained and used within the first 90 seconds of the submission.
\end{itemize}

\begin{figure*}[htb!]
    \centering
    \includegraphics[width=.9\linewidth]{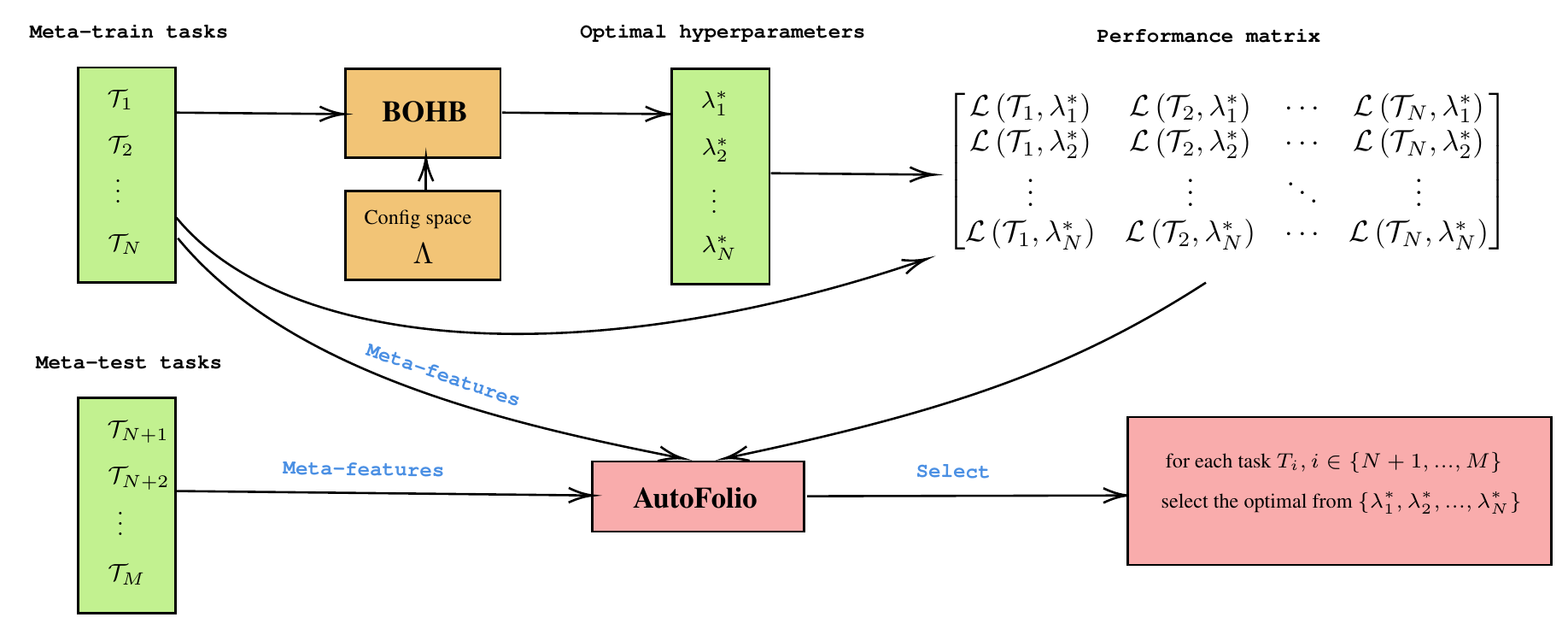}
    \caption{{\bf Workflow of \participant{automl\_freiburg}}. The approach first optimizes the hyperparameter configuration (including choices for training, input pipeline, and architecture) for every task (dataset) in our meta-training set using BOHB~\cite{falkner_bohb_nodate}. Afterwards, for each dataset $i$, the best found configuration $\lambda_i^*$ is evaluated on the other datasets $j\in\{1,2,..., N \}$, $j\neq i$ to build the performance matrix (configurations $\times$ datasets). 
    For training and configuring the meta-selection model based on performance matrix and the meta-features of the corresponding tasks, the approach uses AutoFolio~\cite{lindauer_autofolio_2015}.
    At meta-test time, the model fitted by AutoFolio uses the meta-features of the test tasks in order to select a well-performing configuration.}
    \label{fig:asc}
\end{figure*}

After the extension of the configuration space, they optimized the hyperparameters with BOHB~\cite{falkner_bohb_nodate} across 300 evaluation runs with a time budget of 300 seconds on eight different datasets (Chucky~\cite{cifar}, Hammer \cite{hammer}, Munster \cite{mnist}, caltech\_birds2010 \cite{caltechbirds2010}, cifar100 \cite{cifar}, cifar10 \cite{cifar}, colorectal\_histology \cite{colorectalhistology} and eurosat \cite{eurosat}). 
These eight datasets were chosen from meta-training data to lead to a portfolio of complementary configurations~\cite{xu-aaai10a,feurer-automl18a}.
Additionally, they added a robust configuration to the portfolio of configurations that performed best on average across the eight datasets. Then, they evaluated each configuration of the portfolio for 600 seconds on all 21 image datasets they had collected. In addition, they searched for a tenth configuration (again with BOHB), called the \emph{generalist}, that they optimized for the average improvement across all datasets relative to the already observed ALC scores.
In the end, the meta-train-data consisted of the ALC performance matrix (portfolio configurations $\times$ datasets) and the meta-features from the 21 datasets. These meta-features consisted of the image resolution, number of classes, number of training and test samples and the sequence length (number of video frames, i.e. 1 for image datasets). In addition, they studied the importance of the meta features for the meta-learner, and selected an appropriate subset. 
To optimize the portfolio further, they applied a greedy submodular optimization~\cite{xu-rcra11a,feurer-automl18a} to minimize the chance of wrong predictions in the online phase.
Based on this data, they trained a cost-sensitive meta-model~\cite{xu_satzilla2012_2012} with AutoFolio~\cite{lindauer_autofolio_2015}, which applies algorithm configuration based on SMAC~\cite{hutter-lion11a,smac-2017} to efficiently optimize the hyperparameters of the meta-learner. Since the meta-learning dataset was rather small, HPO for the meta-learner could be done within a few seconds.  Lastly, they deployed the learned AutoFolio model and the identified configurations into the initialization function of the winner's solution code. The workflow of this approach is shown in Figure \ref{fig:asc}.

\end{document}